\documentclass{article}

\usepackage[nonatbib,final]{neurips_2024}
\usepackage[numbers]{natbib}
\usepackage{neurips_2024}

\usepackage[utf8]{inputenc} % allow utf-8 input
\usepackage[T1]{fontenc}    % use 8-bit T1 fonts
\usepackage{hyperref}       % hyperlinks
\usepackage{url}            % simple URL typesetting
\usepackage{booktabs}       % professional-quality tables
\usepackage{amsfonts}       % blackboard math symbols
\usepackage{nicefrac}       % compact symbols for 1/2, etc.
\usepackage{microtype}      % microtypography

\usepackage{subcaption}
\usepackage{wrapfig}
\usepackage{multirow}
\usepackage{makecell}
\usepackage{xspace}
\usepackage{enumerate}
\usepackage{amsmath}
\usepackage{listings}
\usepackage[vlined, ruled]{algorithm2e}
\usepackage{algorithmic}
\usepackage{amsfonts,amssymb}
\usepackage{mathrsfs}
\usepackage{subcaption}
\usepackage{natbib}
\setcitestyle{numbers,square}
\usepackage{graphicx}
\usepackage{cleveref}

\usepackage{paralist, tabularx}

\usepackage{inconsolata}
\usepackage{pgfplots}
\usepackage{booktabs}
\usepackage{epigraph}
\usepackage{float}
\usepackage[normalem]{ulem}
\useunder{\uline}{\ul}{}
\usepackage[framemethod=TikZ]{mdframed}

\usepackage{framed}
\usepackage{mathtools}
\usepackage{array}
\usepackage{amsthm}
\usepackage{verbatim} 
\usepackage{bbm}
\usepackage{commath}
\usepackage{amsbsy}
\usepackage[shortlabels]{enumitem}
\usepackage{tcolorbox}
\usepackage{bbding}
\usepackage{threeparttable}
\usepackage{microtype}
\usepackage{hyperref}
\usepackage{url}
\usepackage{lipsum}
\usepackage{minitoc}
\usepackage{tocloft}

\usepackage{xcolor}
\definecolor{blue-violet}{rgb}{0.54, 0.17, 0.89}
\hypersetup{colorlinks=true, linkcolor=blue-violet, citecolor=blue, urlcolor=blue-violet}
\newcommand{\vpara}[1]{\vspace{0.05in}\noindent \textbf{#1 }}
\newcommand{\ipara}[1]{\vspace{0.03in}\noindent \textit{#1 }}

\usepackage{xr}
\newmdtheoremenv[
  backgroundcolor=gray!8,
  outerlinecolor=black,
  innertopmargin=\topskip,
  splittopskip=\topskip,
  ntheorem=true,
  roundcorner=2pt
]{findings}{Findings}

\title{Training Compute-Optimal Protein Language Models}

\author{
Xingyi Cheng$^{1\ast}$, 
Bo Chen$^{2\ast\dagger}$,
Pan Li$^{1}$,
Jing Gong$^{1}$,
Jie Tang$^{2}$,  Le Song$^{1,3}$\\
$^1$\textmd{BioMap Research}
$^2$\textmd{Tsinghua University}
$^3$\textmd{MBZUAI}\\
 \texttt{derrickzy@gmail.com} \\
 \normalsize\rule{0pt}{1em}Code:~\url{https://github.com/cxysteven/ScalingProteinLM}\\
}

\begin{document}
\doparttoc
\faketableofcontents
\maketitle

\renewcommand{\thefootnote}{\fnsymbol{footnote}}
    \footnotetext[1]{XC and BC contributed equally.}
    \footnotetext[2] {Work done while interned at BioMap.}

\begin{abstract}
%Mention mix training
%mention data intensity, nlp,code,cv
We explore optimally training protein language models, an area of significant interest in biological research where guidance on best practices is limited.
Most models are trained with extensive compute resources until performance gains plateau, focusing primarily on increasing model sizes rather than optimizing the efficient compute frontier that balances performance and compute budgets.
Our investigation is grounded in a massive dataset consisting of 939 million protein sequences. 
We trained over 300 models ranging from 3.5 million to 10.7 billion parameters on 5 to 200 billion unique tokens, to investigate the relations between model sizes, training token numbers, and objectives.
First, we observed the effect of diminishing returns for the Causal Language Model~(CLM) and that of overfitting for the Masked Language Model~(MLM) when repeating the commonly used Uniref database. To address this, we included metagenomic protein sequences in the training set to increase the diversity and avoid the plateau or overfitting effects. 
Second, we obtained the scaling laws of CLM and MLM on Transformer, tailored to the specific characteristics of protein sequence data. 
Third, we observe a transfer scaling phenomenon from CLM to MLM, further demonstrating the effectiveness of transfer through scaling behaviors based on estimated Effectively Transferred Tokens.
Finally, to validate our scaling laws, we compare the large-scale versions of ESM-2 and PROGEN2 on downstream tasks, encompassing evaluations of protein generation as well as structure- and function-related tasks, all within less or equivalent pre-training compute budgets.
%We release all models at \url{https://github.com/cxysteven/ScalingProteinLM}.

%Our findings highlight the importance of data information density in model scalability, offering insights into effective training approaches for protein language models.
%\cxy{More number details.}
%
%Second, by optimally training，Beat ESM, PROGEN2

%Finally, Give fix compute, how to allocation the two objectives.

%We explore optimal training strategies for protein language models, a crucial yet under-researched area in protein research. By compiling a dataset of 930 million protein sequences and training around 200 models ranging from 3.5 million to 7 billion parameters on 5 to 200 billion tokens, we identify key challenges in learning efficiency as models scale. Our analysis reveals distinct scaling behaviors between Masked Language Model (MLM) and Causal Language Model (CLM) optimization objectives, leading us to propose a novel allocation strategy that optimizes model size and training token volume for protein sequence data. This study highlights the importance of data characteristics in model scalability and offers insights into efficient training approaches for protein language models, with broader implications for biological research.
\end{abstract}

\section{Introduction}
%参考Unified router
Scaling up transformer-based models has become a guiding principle for enhancing model performance across broad domains, particularly in Natural Language Processing~(NLP)~\cite{anil2023palm, gpt3, du2022glam, rae2021scaling, touvron2023llama1} and Computer Vision~(CV)~\cite{vision32023scaling,vision2021scaling,vision12022scaling}. 
In recent years, large transformer-based Protein Language Models~(PLMs) such as PROGEN familiy~\cite{madani2020PROGEN,nijkamp2023PROGEN2}, ESM familiy~\cite{rives2021biological,lin2023evolutionary} and xTrimoPGLM~\cite{chen2024xtrimopglm} have also been developed, which leads to significant improvements over model performance on complex downstream tasks~\cite{elnaggar2023ankh, li2024feature}.
Current language models utilize two main training objectives to encode sequence information: the BERT-like~\cite{BERT} Masked Language Model (MLM) and the GPT-like Causal Language Model (CLM)~\cite{brown2020language}, each applied either separately or in a unified fashion.
%And mix objectives, such as xTrimoPGLM~\cite{chen2024xtrimopglm} simutanenously adopts two objective training with huge 100B parameres.
A common understanding is that 
bi-directionally MLM excels in sample efficiency and shows enhanced performance in downstream task fine-tuning. This is particularly true in tasks that emphasize understanding complex patterns, making it a prevalent learning objective in modeling protein sequences\footnote{Appendix~\ref{app:mlm_vs_clm} also compared CLM and MLM on the protein contact prediction task through fine-tuning and freeze probing, with MLM demonstrating superior performance relative to CLM.}
~\cite{ESMfold, chen2024xtrimopglm}.
On the other hand, uni-directional CLM, due to its sequential generation ability, is better suited for generating more coherent and realistic sequences compared to MLM~\cite{dauparas2022robust,nijkamp2023PROGEN2,qiu2024instructplm}. 

%Although it is rarely used in Large (natural) Language Model~(LLM) nowadays,

% It seems that scaling these models of the two objectives allows for the prediction of atomic-level structures~\cite{ESMfold, chen2024xtrimopglm} and generates novel protein sequence candidates~\cite{dauparas2022robust,nijkamp2023PROGEN2,qiu2024instructplm}. 

However, training large protein language models (PLMs) are computational-intensive, and strategies for optimally allocating compute budgets for training PLMs are relatively underexplored, with \textit{most efforts focusing on scaling model parameters based on a fixed set of training tokens to achieve performance improvements}. A key insight~\cite{hoffmann2022training,kaplan2020scaling,tay2021scale} is that large models should not be trained to their lowest possible loss to optimize computing; instead, models and data should be scaled proportionally based on available compute budgets. 
These scaling laws are broadly found in natural language models~\cite{kaplan2020scaling,hoffmann2022training,henighan2020scaling,modality2023scaling,muennighoff2024scaling,tay2022transcending,clark2022unified,zhang2024scaling}.
But their applicability has not been validated within biological datasets, such as the primary structures of proteins, which are composed of amino acid sequences forming protein chains.  
Unlike natural languages, protein sequences are scientific data that are precisely represented using a vocabulary of 20 amino acids, with very little redundancy and are not as semantically smooth. Thus, we consider such data as a distinct modality and ask the question: \textit{What are the scaling behaviors for MLM and CLM in protein language modeling?}

We focus on the best practices, which include revisiting datasets, optimization objectives, and model parameters as key factors. Our goal is to investigate an optimal training scheme for protein language models given predetermined compute budgets. Our core findings are as follows:
\begin{itemize}
  \item We revisited the protein sequence data used for training PLMs and collected a dataset of 194 billion unique tokens on 939M unique sequences from publicly available sources to address the issue of overfitting and perform plateau in protein language modeling. % thereby clarifying previously opaque aspects of protein sequence data.

%    \item The scaling behaviors of both Masked Language Models (MLM) and Causal Language Models (CLM) are described by a power-law. Significant differences were observed between them. For MLMs, the scaling exponent coefficient for computational power and model parameters is \textasciitilde0.77, whereas for CLMs, it is \textasciitilde0.58, which is closer to the power-law observed in natural language or code. 
%    In other words, a \textasciitilde10$\times$ increase in computational power results in approximately a \textasciitilde6$\times$ increase in MLM model size and only a 70\% increase in data. Conversely, for CLMs, the model size increases by 3.8$\times$, and training tokens grows by \textasciitilde2.64$\times$.

  \item We find that, in both MLM and CLM,  training data scales sublinearly in the model sizes but follow distinct power-laws. 
  % MLM scales with a compute exponent of approximately 0.77. 
  In other words, a 10$\times$ increase in compute leads to a 6$\times$ increase in MLM model size and a 70\% increase in data, versus a 4$\times$ increase in CLM model size and a 3$\times$ increase in training tokens.

  \item We also find that models trained with CLM can be transferred to MLM. When given a predetermined amount of computation, and one wants to obtain both a CLM and a MLM model, there is a trade-off in allocating the training token to each model to jointly optimize the performance of the two. Interestingly, the allocation for CLM pre-training was determined by the scaling law of CLM and MLM, and the Effectively Transferred Tokens $D_t$ from CLM to MLM. 
  Furthermore, we verify this method experimentally using a 470M model and fine-tuning on downstream tasks.
  
  %\item With the same computational resources measured in FLOPs, a strategic combination of effective CLM and MLM pre-training was found to surpass the baseline established by MLMs trained from scratch.
  \item Building on our scaling strategies, we re-allocate   of model size and training tokens under the compute budgets of established PROGEN2-xlarge and ESM-2 (3B) setups. Consequently, with the same compute budgets, we trained two corresponding models, one with 7.2B parameters and the other with 10.7B parameters, which exhibited enhanced performance in a diverse range of downstream tasks.   %\cxy{rewrite, more detail numbers}

\end{itemize}
%Specifically, it seeks to determine how to optimally balance model size and data size given a fixed computational budget and training objectives, enabling models to anticipate returns on expanded resources.

% we run how many model, and range
%We first surveys datasets and, based on current scaling laws, assumes that repeating training datasets can detrimentally degrade the model perforamnce. Therefore, we scale the datasets to ensure that the model computes only one epoch given the available FLOPs. 

\vspace{-3mm}
\section{Scaling up data}
\vspace{-5pt}
First, we explore the effects of training PLMs across multiple epochs under token scarcity conditions. 
We then introduce a dataset, UniMeta200B, used throughout this work.
This dataset enhancement alleviates the challenge of insufficient training for protein language models. 
\vspace{-5pt}
\subsection{A Data-hungry Observation}
\vspace{-5pt}
\begin{figure}[ht]
\centering
    \includegraphics[width=0.95\textwidth]{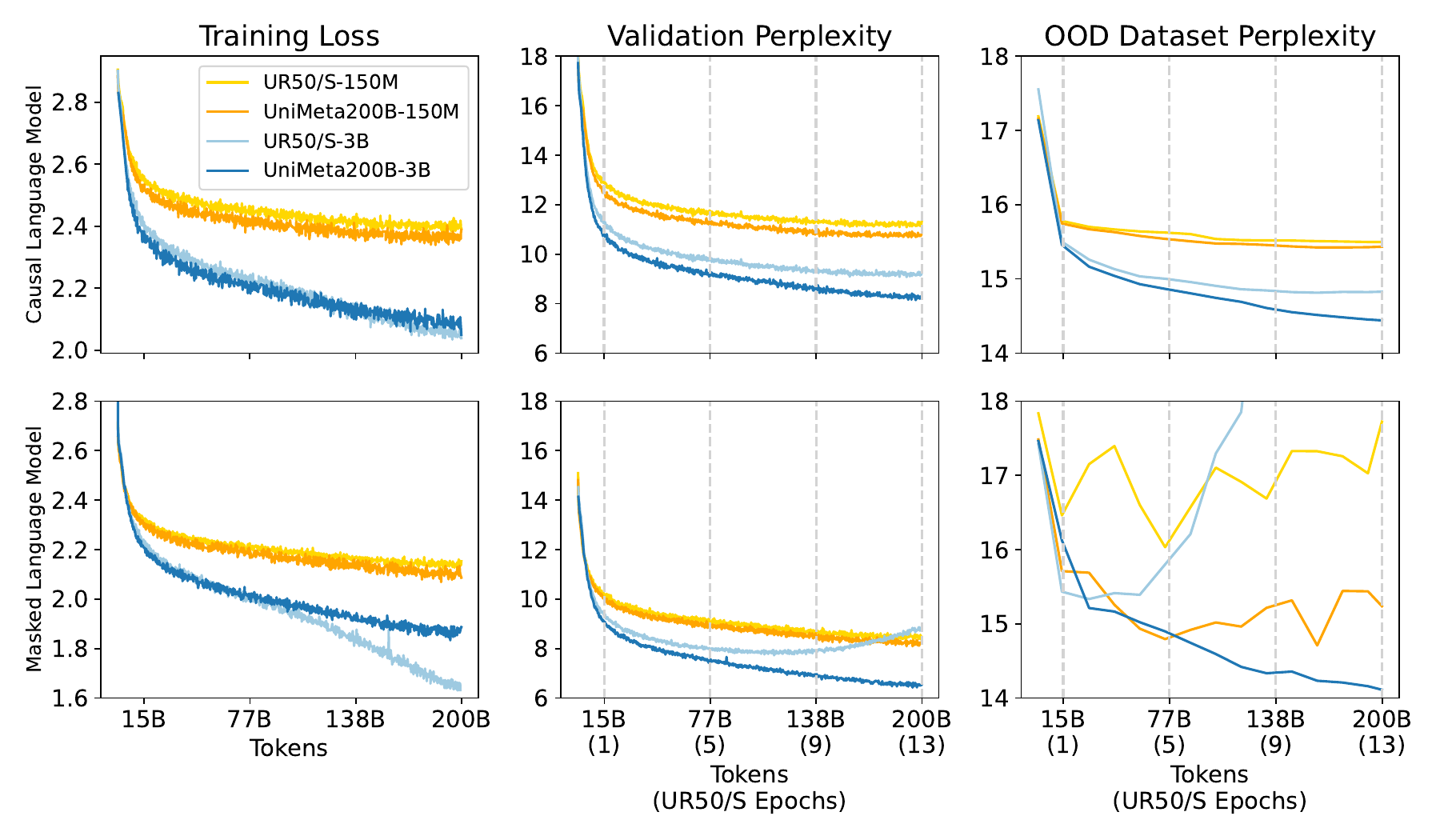}\vspace{-3mm}
	\caption{\textbf{Learning curves for UR50/S and UniMeta200B.} Training loss and validation PPL, OOD test PPL, were tracked over 200 billion training tokens for both the 150M and 3B models. As we scaled the model from 150M to 3B, we observed diminishing returns on CLM~(First line) and a tendency to overfit on MLM~(Second line) when repeating the Uniref50~(UR50/S) dataset. We totally evaluate 3 repeating methods on MLM 3B models, all of which present overfitting~(see Appendix~\ref{app:repeat}).  % By enlarging the metagenome dataset, we can achieve an increase in sample efficiency.\cxy{ color and spike}
    }
    \label{fig::overfit}
   \vspace{-20pt}
\end{figure}

%Using the UniParc database with 250 million protein sequences, research on ESM~\cite{rives2021biological} demonstrates that the diverse datasets UR50/D and UR50/S, each with 45M and 65M unique sequences, outperform the less diverse Uniref100 in Perplexity (PPL) on a MLM model of \textasciitilde670M parameters. These datasets respectively contain \textasciitilde15B and \textasciitilde20B unique amino acid tokens. The ESM-2 family models, ranging from 150M to 15B in size, undergo extensive training with nearly 1 trillion tokens over 45 epochs on the UR50/D dataset. 
%However, contemporary LLMs are typically trained for one or a few epochs~\cite{komatsuzaki2019one,hoffmann2022training,touvron2023llama1,touvron2023llama2,schaper2012repeat}.
%Heavy data repetitions with insufficient unique token counts can degrade performance and hinder model scaling~\cite{raffel2020exploring,hernandez2022scaling,muennighoff2024scaling,hernandez2022scaling}. Such limitations highlight the importance of training large-scale language models on datasets rich in unique tokens to ensure robust performance across diverse applications.

Using the UniParc database with 250 million protein sequences, research on ESM~\cite{rives2021biological} shows that the datasets UR50/S and UR50/D, with 45M and 65M unique sequences respectively, outperform Uniref100 in perplexity (PPL) on a \textasciitilde670M parameter MLM model. These datasets contain \textasciitilde15B and \textasciitilde20B unique amino acid tokens. The ESM-2 family models, ranging from 150M to 15B parameters, are trained extensively with nearly 1 trillion tokens over 45 epochs on the UR50/D dataset. In observing the scaling of ESM-2 models, it becomes apparent that increasing model size to 15B parameters from 3B shows marginal improvement. 
On the other hand, contemporary LLMs are often trained for only one or a few epochs~\cite{ komatsuzaki2019one,hoffmann2022training,touvron2023llama1,touvron2023llama2,brown2020language}. The repetition of data with limited unique tokens has diminishing returns and hinders scaling model size~\cite{raffel2020exploring,hernandez2022scaling,muennighoff2024scaling,schaper2012repeat}. This underscores the importance of using rich datasets for training large-scale language models to ensure robust performance across applications. 
We evaluated models with 150M and 3B parameters on the UR50/S dataset, trained on 200B tokens, as shown in Figure~\ref{fig::overfit}. We focus on the Independent and Identically Distributed~(IID) validation and  Out-Of-Distribution~(OOD) test PPL, which measures the model's randomness in amino acid selection. 
For our OOD dataset, we utilized the MMseqs2 tool~\cite{steinegger2017mmseqs2} to conduct searches within the UniRef90 database for sequences post-training dataset timestamp, retaining those with \textit{no detectable} identity. From these, a random sample of 3,000 sequences was selected to constitute the OOD dataset.
Notably, we do not adopt dropout regularization, a practice that often reduces model capacity and is infrequently used in contemporary LLMs~\cite{komatsuzaki2019one}. This choice is consistent with recent LLM configuration findings~\cite{huggingface_llama2}, including ESM-2~\cite{lin2023evolutionary}.

The results show the 150M model lacks good generalization while increasing to a 3B model resulted in diminishing returns for CLM and severe overfitting for MLM.  
% It suggests the CLM tasks are more challenging than MLM. 
Principally, the bidirectional self-attention mechanisms used in MLM have a higher capacity to overfit compared to the unidirectional self-attention used in CLM. This is because MLM can utilize the entire context surrounding a masked token, leading to faster memorization of the training data.

\begin{wraptable}{r}{0.60\textwidth} 
\centering
\small
\caption{\textbf{The Pre-training data}, aggregates various public sources and specifies sampling proportions for a single epoch of training on 194 billion unique amino acids.}
\label{tab::unimeta}
% \begin{small}
% \begin{sc}
\begin{tabular}{lccc}
\toprule
 Datasets & Prot. Seq.  &  Tokens~(AAs) & Samp. Prop. \\
 \midrule
 Uniref50/S & 54M & 15.2B &   8.5\%  \\
 Uniref90/50 & 102M &  37.8B &  19.5\%  \\ 
 ColabFoldDB$_c$ & 208M & 37.7B &  19.5\%  \\
 ColabFoldDB$_m$ & 575M & 103B &  52.5\%   \\
 Total & 939M & 194B &  -  \\
\bottomrule
\end{tabular}
\end{wraptable}

%ESM: Transformers trained on the two high-diversity datasets, UR50/S and UR50/D, improve generalization over the UR100low-diversity dataset. The best Transformer trained on the most diverse and dense dataset reaches an PPL of 8.46, 
\vspace{-10pt}
\subsection{Expanding Diversified Metagenomic Data}
\vspace{-5pt}
To tackle the challenge of data scarcity, we leveraged the ColabFoldDB database~\cite{mirdita2022colAbFold}, which focuses on metagenomic data sources such as BFD~\cite{bfdmmseq}, MGnify~\cite{mitchell2020mgnify}, and specific eukaryotic and viral datasets including SMAG~\cite{delmont2022functional}, MetaEuk~\cite{levy2020metaeuk}, TOPAZ~\cite{alexander2021eukaryotic}, MGV~\cite{nayfach2021metagenomic}, and GPD~\cite{camarillo2021massive}. We applied a stringent deduplication process with a maximum similarity threshold of 0.3 to preserve the diversity of the protein universe. 
%Compared with the previous largest pre-training data set, UniRef90, it has increased by an order of magnitude.
%as detailed in Table~\ref{tab::unimeta}.
Given that the Uniref90 dataset has proven most effective for pre-training across various Uniref clustering levels per ESM-1v~\cite{meier2021language}, we incorporated Uniref90/50~(Before 2022-12), which includes incremental data relative to Uniref50/S representatives. 
ColabFoldDB$_c$ and ColabFoldDB$_m$ play dominant roles within the dataset, corresponding to cluster representatives and members, respectively. To ensure uniformity during training, we allocate weights within each batch to allow each amino acid token to be evenly processed through the model.
This dataset, termed UniMeta200B, contains 939 million unique protein sequences and 194 billion amino acids, which is an order of magnitude larger than UR50/D.
%we prioritize samples from ColabFold in each batch according to their proportional representation in the dataset.
We observed significant improvements in the OOD test set and a consistent learning curve on the IID validation subset extracted from the training set~(Figure~\ref{fig::overfit}). 
These enhancements not only ensure a controlled diversity to maintain sample efficiency but also significantly increase the quantity and uniformity of data, facilitating model scaling.~\footnote{Appendix~\ref{app:data_quality} compare the training performed separately on two datasets, and we find that the ColabFoldDB does not affect downstream results.}

\begin{findings}
Scaling the model from 150M to 3B, we noted diminishing returns for CLM and an overfitting tendency for MLM when repeating the UR50/S dataset.
%\end{findings}
%\begin{findings}
The proposed Expanding Diversified Metagenomic Data (UniMeta200B) addresses these problems.
\end{findings}

\section{Parameters and Datasize Optimal Allocation}

\begin{figure}[!ht]
	\centering
	\includegraphics[width=0.85\textwidth]{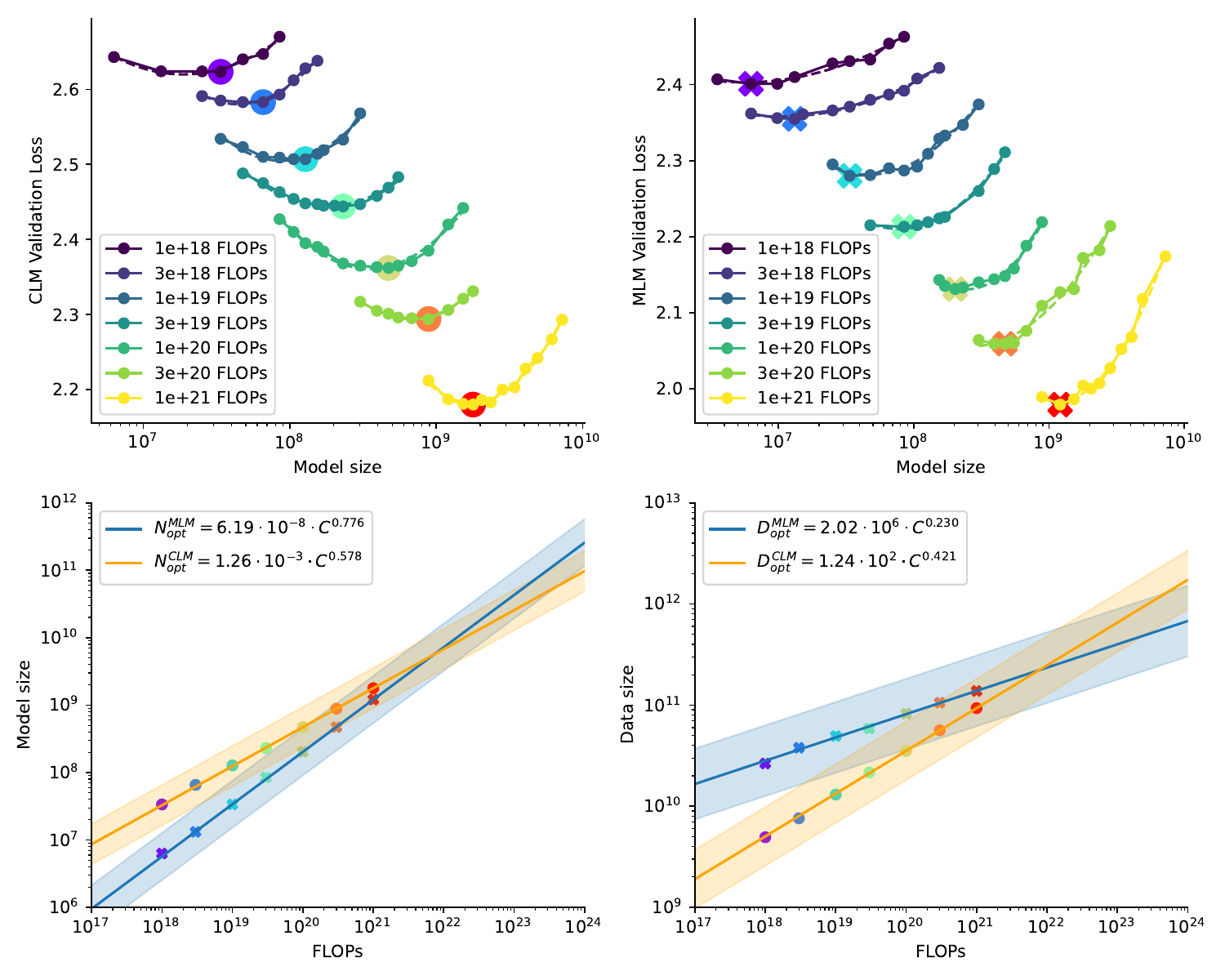}
    \vspace{-3mm}
	\caption{\textbf{IsoFLOPs curves and parametric fit for CLM and MLM.} We selected training tokens to ensure a uniform final FLOP count for different model sizes. The lowest loss of each curve revealed an optimal model size for a FLOP budget ~(\textbf{above}). We use these rainbow points at the valley to plot the efficient frontier for estimating the optimal model size and training tokens for scaling models~(\textbf{below}). The interval range was estimated by model points with similar loss.}
    \label{fig::scaling_law}
    \vspace{-5pt}
\end{figure}

%\cxy{ rainbow color means representative points, in the interval range will getsimilar loss of it. the interval error is [-0.3,-0.3]

%low right: derived from low left by D = FLOPs/6D. demonstrating shadow is follow ISOLoss(app) calculated error,这个模型size的误差表示在某个FLOPs上,他们会获得类似的loss, 
%    TODO: add some small model. It may demonstrate that mlm training use small model for multiple epoch will not hurt performance. We constrain that Training tokens upper bound for one epoch.
%    How many total models we run ?

In this section, we propose a scaling law for protein sequences with MLM and CLM objectives, aiming at optimally balancing model size and data size under a fixed compute budget to improve efficiency on expanded resources.
\vspace{-3mm}
\subsection{Scaling laws for CLM and MLM}
\label{sec::scaling_law}

%\cxy{1. demonstrate the approach, chinchilla approach1 and approach2, full learning schedule. 解释：我们没有使用更加复杂的形式来拟合数据，考虑到数据的多样性以及1 epoch training,可以假设在无穷大的参数和无穷多的数据上进行训练，basic scaling law  L(N) 和 L(D)大体可以决定L(N,D).   The D(N)可以直接eliminate C来得到. 
%2. model fit methods
%}

\begin{wraptable}{r}{0.60\textwidth} 
\centering
\small
\vspace{-20pt}
\caption{Coefficient of Equation~\ref{eq:c_law}.}
\begin{tabular}{@{}lcccc@{}}
\toprule
Parameter & $\alpha$ & $\beta$ & $A$  & $B$  \\ 
\midrule
CLM & $0.578$     & $0.422$      &  $1.26 \times 10^{-3}$  &  $1.23 \times 10^{2}$  \\
MLM &   $0.776$       & $0.230$            &  $6.19 \times 10^{-8}$    & $2.02 \times 10^{6}$ \\
\bottomrule
%\vspace{-40pt}
\end{tabular}
\label{tab:c_law}
\end{wraptable}

We first fit our models in the form of a fundamental power-law based on the existing work~\cite{kaplan2020scaling, hoffmann2022training, henighan2020scaling,robert2022unconstrained,modality2023scaling,clark2022unified, zhang2024scaling} in the field of LLMs.
Specifically, given a fixed FLOPs formula of $ C = 6 \times N \times D$, where $N$ represents the number of forward-activated non-embedding parameters, and $D$ is the number of training tokens, how should one navigate the trade-off between model size and the number of training tokens?  The model parameters $N$ and data size $D$ can be directly fit with a simple power-law:
\begin{equation}
    N(C) = A \times C^\alpha, \quad D(C) = B \times C^\beta 
%    \\
%    \quad \alpha_{\text{MLM}} \sim 0.578 \quad \beta{\text{MLM}} \sim 0.422 \quad A_{\text{MLM}} \sim 1.26 \times 10^{-3} \ quad B_{\text{MLM}} \sim 1.23 \times 10^2
\label{eq:c_law}
\end{equation}

We employed the IsoFLOPs profiling approach~\cite{hoffmann2022training,bi2024deepseek}, setting 7 distinct training FLOP counts ranging from $1 \times 10^{18}$ to $1 \times 10^{21}$. For each FLOP count, we selected models from a pool of candidates~(see Appendix~\ref{app::modelsize}). Models were excluded if the estimated data size ~($C/(6*N)$) resulted in more than 200B tokens or if the training steps were fewer than 20K. Ultimately, approximately 260 models were used for fitting. We considered the final validation loss for each model to ensure that every model completed a full cosine cycle with $10\times$ learning rate decay.
%\footnote{different learning rate schedules with the same number of steps can significantly impact performance. See Chinchilla~\cite{hoffmann2022training} appendix C}.
For each fixed FLOP count, we employ smoothed loss to determine the optimal model size with the smallest loss~(Figure~\ref{fig::scaling_law}~(above)). Subsequently, we use Equation~\ref{eq:c_law} and apply the \texttt{least\_squares} method to fit the model. %specifically utilizing \texttt{scipy.optimize.curve\_fit}.

Given the minimal variations in the final loss among a set of $(N, D)$ configurations, we classify these configurations as operating under "IsoLoss" conditions~(see Appendix~\ref{app::isoloss} Figure~\ref{fig::isoloss}), considered optimal for training. In Figure~\ref{fig::scaling_law}~(below), we illustrate an efficient frontier interval that demonstrates permissible fluctuations in model size and dataset size at a specific FLOP count, while still achieving nearly identical losses. The variation in loss is quantified at 0.25 on a logarithmic scale with a base of 10. This indicates that within this FLOP counts, the model size can be adjusted within a range, increasing up to 80\% or decreasing up to 40\% without repeating data, to maintain a loss variation within 0.01. 

%We find different results~(Table~\ref{tab:c_law}) in the rate of proportional growth between the MLM model and the number of training tokens as compared to the CLM.  Firstly, prior to the intersection point around 1e-22, as illustrated in Figure~\ref{fig:scaling_law}(left below), the model size of MLM is generally smaller than that of CLM. Secondly, both models demonstrate that, with an increase in the compute budgets, the growth rate of the model size should exceed that of the training tokens, the growth rate for the MLM model is significantly higher than that for the CLM. 
%\cxy{likely because the MLM are more sample efficient than CLM?}
%For instance, if the compute budget is increased by $10\times$, the size of the CLM model should increase by $4\times$ and the training data volume by $3\times$, aligning more closely with proportional scaling. For the MLM, the model size should increase by $6\times$ and the training data size by $1.7\times$.
\vspace{-1mm}
We observe distinct growth rates in the proportional relationship between model size and training tokens for the MLM model compared to the CLM, as detailed in Table~\ref{tab:c_law}. 
Both models demonstrate an increase in the growth of model size that surpasses the growth of training tokens. 
Up to the intersection point around \(1 \times 10^{22}\) (see Figure~\ref{fig::scaling_law}, left below), the model size of MLM tends to be smaller than the CLM, thereafter, the MLM rapidly exceeds that of the CLM. 
Notably, the growth of the MLM's training tokens is greatly lower than that for the CLM, possibly due to MLM's higher sample efficiency.
For instance, if the compute budget is increased by $10\times$, the size of the CLM model should increase by $4\times$ and the training data by $3\times$, aligning more closely with equally proportional scaling. For the MLM, the model size should increase by $6\times$ and the training data size by $1.7\times$.

In exploring the scaling relations of loss, we analyzed various model sizes \(N\), compute budgets \(C\), and training dataset tokens \(D\). These can be described by a similar power-law relation defined as:
\begin{equation}
    L(x) = \beta_x \times x^{\alpha_x}
\label{eq::individual}
\end{equation}
where \(\alpha_x\) is the scaling exponent for different variables. For each FLOP count, we aimed to identify the minimal loss as the fitting target along with the corresponding independent variable \(x\).
%using the least squares method similar to the one previously described.
Table~\ref{tab:l_law} presents these fitting coefficients.
\begin{table}[h]
\vspace{-3mm}
\small
\centering
\caption{Coefficient of Equation~\ref{eq::individual} }
\begin{tabular}{@{}lcccccc@{}}
\toprule
Objective & $\alpha_N$ & $\alpha_D$ & $\alpha_C$ & $\beta_N$  & $\beta_D$ & $\beta_C$  \\ 
\midrule
CLM & $-0.037$     & $-0.051$  &  $-0.027$   &  $4.835$  &  $7.904$ &  $8.251$ \\
%MLM &   $ -0.041$       & $-0.113$ &    $-0.034$       &  $4.710$   & $35.647$ & $10.125$ \\
MLM &   $ -0.040$       & $-0.120$ &    $-0.034$       &  $4.530$   & $42.614$ & $10.125$ \\
%MLM &   $ -0.040$       & $-0.119$ &    $-0.034$       &  $4.535$   & $42.33$ & $10.125$ \\

\bottomrule
\end{tabular}
\label{tab:l_law}
\end{table}

%\cxy{the scaling exponent is very sensitive to the results, for example , change the 1e-4 precision ,will have 1e9 data or modelszie varying.}

Based on the coefficients obtained from the fitting described above, we can establish the relationship between \(D\) and \(N\) by eliminating \(L\). The relationship is expressed by the following equation:
\begin{equation}
%D(N) = B \times \left(\frac{N}{A}\right)^{\frac{\beta}{\alpha}} 
D(N) = \left( \frac{\beta_N}{\beta_D} \right)^{\frac{1}{\alpha_D}} \times N^{\frac{\alpha_N}{\alpha_D}}
%D(N) = \left( \frac{\beta_N}{\beta_D} \right)^{\frac{1}{\alpha_D}} \times N^{\frac{\alpha_N}{\alpha_D}}
\end{equation}
By substituting the learned coefficients into this formula, we can derive $D^{\text{opt}}_{\text{MLM}}$ and $D^{\text{opt}}_{\text{CLM}}$ when given $N$. The estimation may be affected when the data exceeds 200 billion or when the quality or quantity of the training dataset changes. 

%Following both individual power-laws, it is possible to integrate two independent scaling laws~(see Appendix~\ref{app::joint}) and allocate two FLOPs within a specified compute budget to train two optimal models if our goal is to simultaneously obtain both optimal training models.
%\begin{wraptable}{r}{0.60\textwidth} 
\begin{wrapfigure}{r}{0.4\textwidth}
    \centering
    \vspace{-50pt} 
    \includegraphics[width=0.40\textwidth]{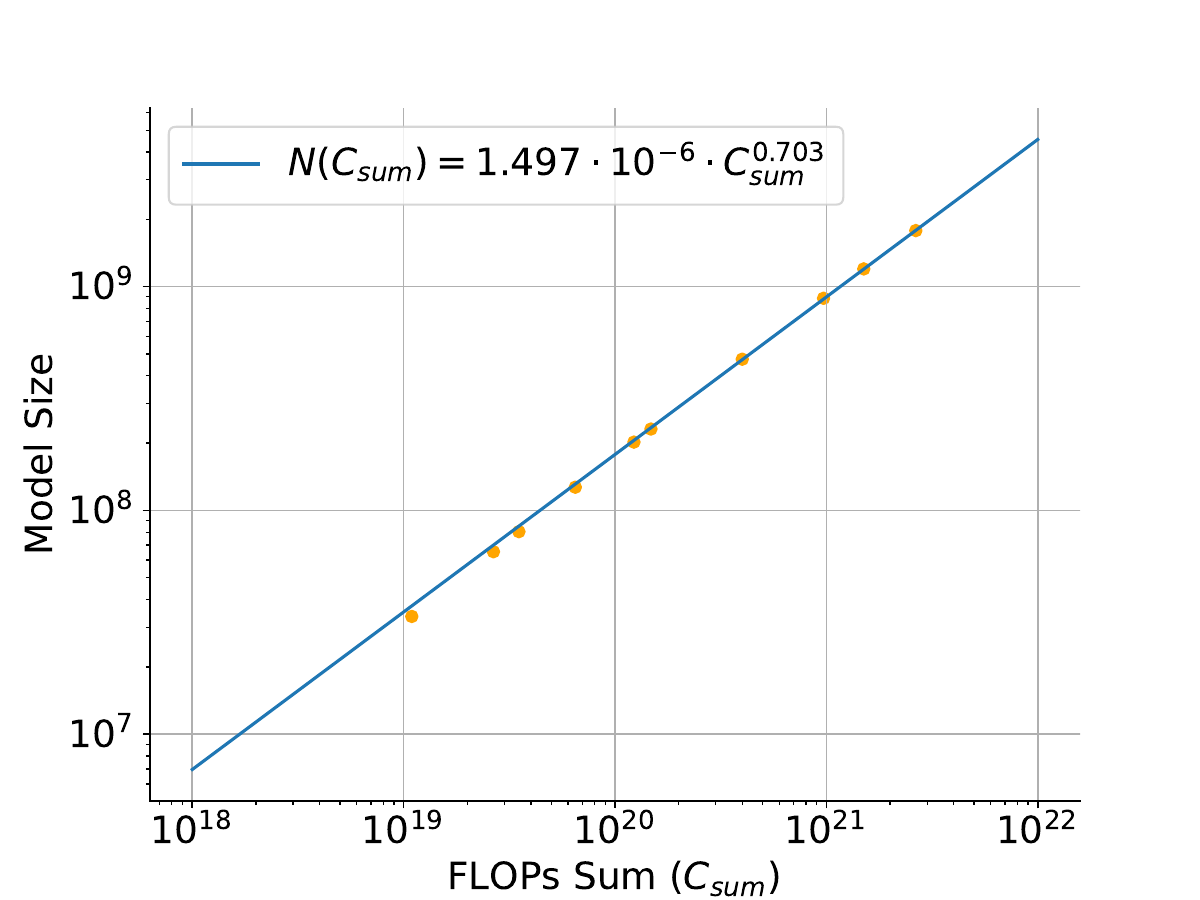}
    \caption{Compute allocation for two objectives with the same model size.}
    \label{fig:joint}
   % \vspace{-20pt} 
\end{wrapfigure}

%\begin{wrapfigure}{r}{0.4\textwidth}
 %   \centering
  %  \vspace{-40pt} 
    %\includegraphics[width=0.45\textwidth]{figures/mix_model_size_vs_flops.pdf}
    %\caption{Compute allocation for two objectives with the same model size.}
    %\label{fig:joint}
   % \vspace{-40pt} 
%\end{wrapfigure}
\vspace{-3mm}
\subsection{Scaling law for training two models}

When our goal is to optimize both CLM and MLM simultaneously, the strategic allocation of compute resources between these two objectives becomes essential. 
To facilitate this, we equalize model parameters across objectives to assess specific compute budgets for dual-objective training. 
Specifically, we seek the compute budgets, \(C_{\text{MLM}}\) and \(C_{\text{CLM}}\), for configurations where the optimal model size is the same, i.e., \(N(C_{\text{MLM}}) = N(C_{\text{CLM}})\). These individual computations are then aggregated to formulate the overall compute budget:

\begin{equation}
C_{\text{sum}}(N) = C_{\text{MLM}}(N) + C_{\text{CLM}}(N) = \left(\frac{6.2 \times 10^{-8}}{N}\right)^{0.776} + \left(\frac{1.25 \times 10^{-3}}{N}\right)^{0.578}
\end{equation}
These two objectives share the same parameter size, their compute budget \(C\) and the number of training tokens \(D\) differ. Thus we further introduce a model-to-ratio \(r(N)\) as \(D_{\text{MLM}}(N) / D_{\text{CLM}}(N)\). We then achieve the relationship between \(N\) and \(C_{\text{sum}}\) by a fitted power-law~({Figure~\ref{fig:joint}) form:

\begin{equation}
\begin{cases}
N(C_{\text{sum}}) \approx 1.497 \times 10^{-6} \times C_{\text{sum}}^{0.703} \\
r(N) \approx 8.449 \times 10^{4} \times N^{-0.392}
\label{eq:joint}
\end{cases}
\end{equation}
The ratio \(r(N)\) informs us about the allocation proportion of training tokens. Specifically, under equal parameters, the data for MLM should exceed that for CLM until a 10B threshold (achieving a 1:1) is reached, after which more training tokens are allocated to CLM.

%Appendix~\ref{app::joint} derives a scheme based on two independent power-laws to guide how to allocate these FLOPs between the MLM and CLM models.
%If our goal is to achieve to two models simultaneously, 
%\cxy{If our goal is that optimize two objective simultaneously, see 
%\newpage
We further find that the scaling behavior of sparse parameter counts in a Mixture of Experts (MoE) model, set with eight experts (see Appendix~\ref{app:moe}), as well as a combined power-law formula used to fit our data (see Appendix~\ref{app:combine}), both exhibit a certain similarity to the scaling behavior we have proposed. 
%We hope that this discovery can be applied in future implementations.

\begin{findings}
 In both CLM and MLM, training data scales sublinearly with model size, following distinct power laws. With an ``infinite'' dataset, where samples are not repeated and training for less one epoch, MLM’s model size grows faster than CLM’s.
    %A 10× increase in compute results in a 6× increase in model size and a 70\% increase in training data.
\end{findings}

% \begin{observation}
% If a reasoning route starting from $p_{k}$ requires more steps to get to the correct answer, then the single-step weighted reward $w_{s_{k}}$ is lower.
% \end{observation}

\section{Transfer Scaling}
\label{sec::optimal_transfer}
\begin{figure}[bht]
	\centering
        \label{fig:scaling_law}
	\includegraphics[width=0.85\textwidth]{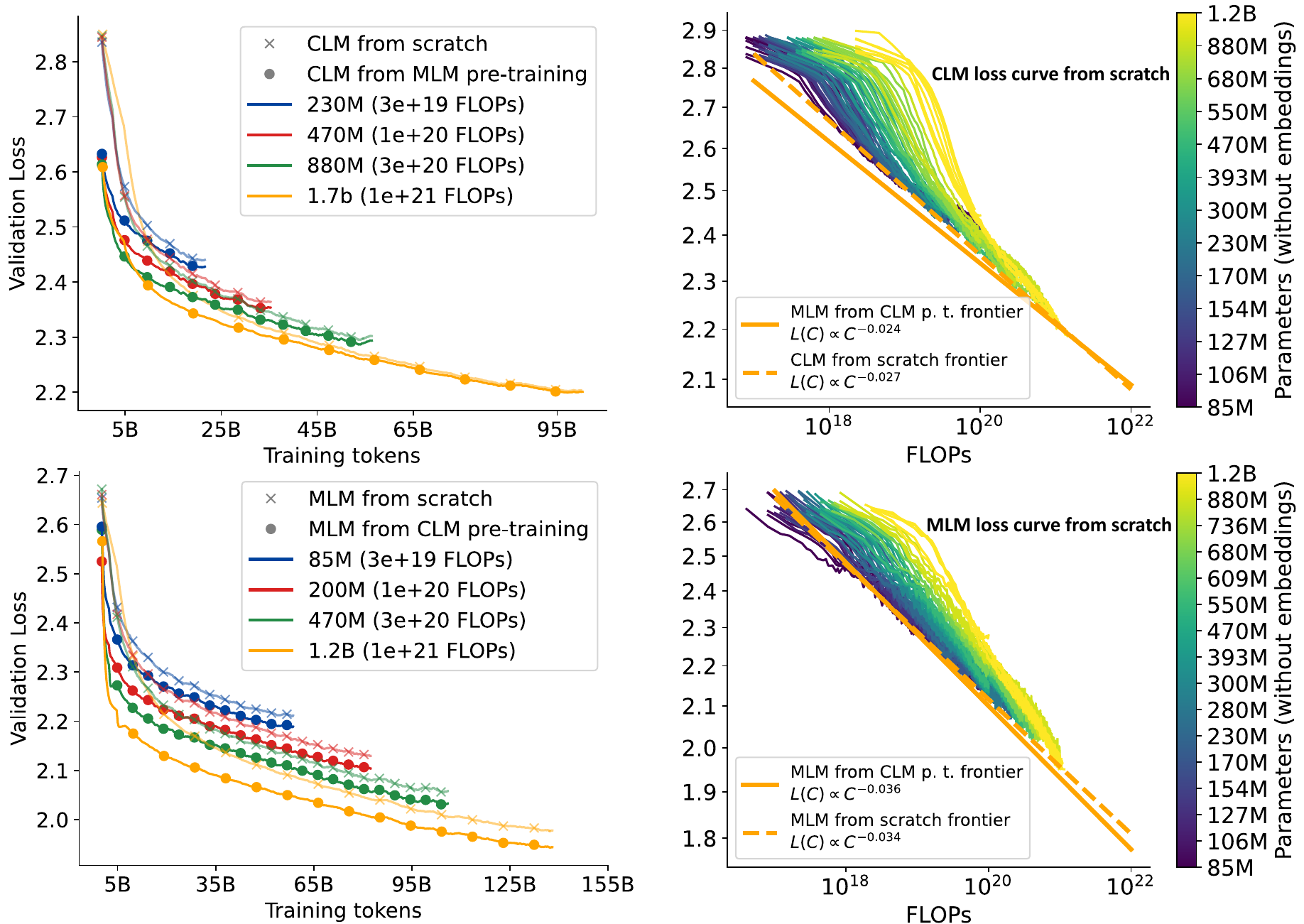}
\vspace{-1mm}
\caption{\textbf{Left:}
   The upper graph compares validation loss of CLM trained from scratch with those transferred from MLM, showing diminishing transfer benefits as model size increases. The lower graph depicts increased benefits for MLM from pre-trained CLM with larger sizes, indicating scale-dependent efficiency gains.
   \textbf{Right:}
    Shows loss curves for CLM and MLM across different FLOPs, emphasizing the efficient frontiers~(or Pareto Frontier) from various transfer strategies. It highlights that the benefits of transferring from CLM to MLM grow with model size, reflecting a scale-dependent synergy between training objectives.}
    \label{fig::transfer}
    %\vspace{-10pt}
\end{figure}

We have outlined two independent scaling laws and how to allocate FLOPs under a fixed budget for training two optimal models, one with MLM and the other with CLM. However, we have not explored the interaction between these objectives. This raises important questions: Can models trained with one objective transferred to one with another objective? Is there a synergistic effect from training two models? Does training order impact the results?

\subsection{Transferability}

We conduct transfer learning experiments on MLM and CLM objectives, selecting eight optimal model sizes based on Equation~\ref{eq:c_law}. These models correspond to four increasing FLOP counts from \(3 \times 10^{19}\) to \(1 \times 10^{21}\) and undergo training from scratch followed by transfer training. Transfer training involves initially training on MLM or CLM, then training on the alternate model for each size.

%We conducted transfer learning experiments on these two objectives, selecting 8 optimal model sizes for MLM and CLM based on Equation~\ref{eq:c_law}, corresponding to 4 different larger FLOP counts (ranging from \(3 \times 10^{19}\) to \(1 \times 10^{21}\)). Each model runs training from scratch and transfer training. Specifically, transfer learning involved initially training on MLM/CLM, followed by training on the corresponding CLM/MLM for that model size.

%We find that optimal pre-training on one objective provided corresponding benefits to the target objective in transfer learning. 
%However, the effect differed between the two methods; starting with CLM and then training MLM, the benefits increased with the model scale. For instance, starting with MLM and then training CLM, the benefits began to diminish. 
%Figure~\ref{fig::transfer} (above left) shows that for a model size of 230M with \(3 \times 10^{19}\) FLOPs, training CLM first and then MLM reduced the loss compared to training from scratch by 0.02, but this benefit approached zero for the 1.7B model.
%Conversely, as seen in Figure~\ref{fig::transfer} (below left) from models ranging from 85M to 1.2B, the transfer benefits gradually increased compared to training from scratch, with validation loss decreasing from 0.045 to 0.025. 
%We speculate that this phenomenon is due to CLM having higher sample efficiency than MLM. CLM calculates all token losses for a single protein sequence, while MLM only has 15\%.

We find that optimal pre-training on one objective benefits the target objective in transfer learning, though effects vary between methods. Starting with CLM and then training MLM, benefits increase with model scale. In contrast, starting with MLM then training CLM sees diminishing benefits. As shown in Figure~\ref{fig::transfer}~(left), for a model size of 230M with \(3 \times 10^{19}\) FLOPs, MLM from CLM pre-training reduces the loss by 0.02 compared to MLM from scratch, however, benefit that nears zero for the 1.7B model. Conversely, for models from 85M to 1.2B, transfer benefits grow with model size, the compared validation loss gap increasing from 0.025 to 0.045. This likely stems from the higher loss utilization rate in MLM; CLM calculates losses for all tokens in a protein sequence, whereas MLM only calculates losses for 15\% of the tokens.
~\footnote{Appendix~\ref{app:mask_ratios} analyzes the mask ratios.
}.

%\cxy{Large model size transfer have seen the double descent phoenmeno.}

We use a power-law to model the transfer scaling law, initially excluding the pre-training FLOPs.
The scaling behavior of transfer learning is modeled by:
\begin{equation}
\label{lc}
L(C_s) = A_s \times C_s^{\alpha_s}, \quad L(C_t) = B_t \times C_t^{\alpha_t}
\end{equation}
where $L(C_t)$ and $L(C_s)$ represent the loss for transfer learning and training from scratch.

\begin{wraptable}{r}{0.50\textwidth} 
\centering
\small
\vspace{-15pt} 
\caption{Coefficients for $L(C_s)$ and $L(C_t)$}
\begin{tabular}{@{}lcccc@{}}
\toprule
Parameter &$A_s$ & $\alpha_s$ & $B_t$ & $\alpha_t$  \\ 
\midrule
MLM & $10.125$     & $-0.034$      &  $11.133$  &  $-0.038$  \\
CLM &   $  8.251$     & $-0.027$            &   $7.191$   & $-0.024$ \\
\bottomrule
\label{tab:trans}
\vspace{-20pt} 
\end{tabular}
\end{wraptable}

Figure~\ref{fig::opt_trans}~(right) shows that the efficient frontier for $L(C_t)$ has shifted relative to $L(C_s)$~(it can be directly obtained from Table~\ref{tab:l_law}, repeated here for convenience.), indicating an improvement. The coefficients from both are shown in Table~\ref{tab:trans}, where we can infer that $C_t \propto C_s^\frac{\alpha_s}{\alpha_t} = C_s^{0.89}$, suggesting that training MLM from scratch with 10$\times$ the compute requires approximately 7.7$\times$ the compute compared to MLM from CLM pre-training.
Another observation is that mixing training objectives in a single batch tends to be detrimental. Detailed results and settings are in Appendix~\ref{app::mix}. The recommended transfer learning schedule involves pre-training CLM before MLM, as mixed training and order swapping show no benefits. 
We speculate that this primarily occurs because our MLM, which focuses solely on recovering corruption tokens, is not causal. If it predicted a middle segment in a left-to-right manner, it could mutually adapt with the context to accelerate training~\cite{wang2022language}.

\begin{findings}
Transferring from MLM to CLM results in diminishing returns.
Conversely, transferring from CLM models to MLM models remains effective as compute scales.
\end{findings}

%As previously, we use a power-law to fit the transfer scaling law. Here, we initially exclude the pre-training FLOPs (to be considered in the next section), and fit the scaling behavior of transfer learning solely through the following formula:

%In Figure~\ref{fig::opt_trans}~(right), the efficient frontier (or Pareto frontier for performance and compute) for $L(C_t)$ has shifted relative to $L(C_s)$, resulting in an improvement. The coefficients for both tables are shown in Table~\ref{tab:trans}.
%Specifically, we could see $C_t \propto C_s^\frac{\alpha_s}{\alpha_t} \sim C_s^{0.89}$, which means, the compute of training MLM from scratch $10\times$, the MLM from pre-trained CLM need $7.7\times$ compute. 

%Additionally, we observe that a random mix of objectives, i.e., a batch containing samples optimized for two different objectives, tends to be detrimental to each other. Detailed results and training settings are provided in Appendix~\ref{app::mix}. In summary, the transfer learning schedule requires pre-training the CLM first, followed by the MLM. Mixed training can harm each other, and swapping the order yields virtually no benefit. 
%We speculate that this is primarily because the MLM we employ is not of a causal mode; it focuses solely on recovering corruption tokens them-selves. If it were to predict a middle segment in a left-to-right behavior, both could mutually adapt to accelerate training~\cite{wang2022language}.

%\begin{table}[h]
%\centering

%\end{tabular}
%\end{table}

\vspace{-3mm}
\subsection{Effectively Transferred Tokens}

\begin{figure}[ht]
	\centering
\includegraphics[width=0.85\textwidth]{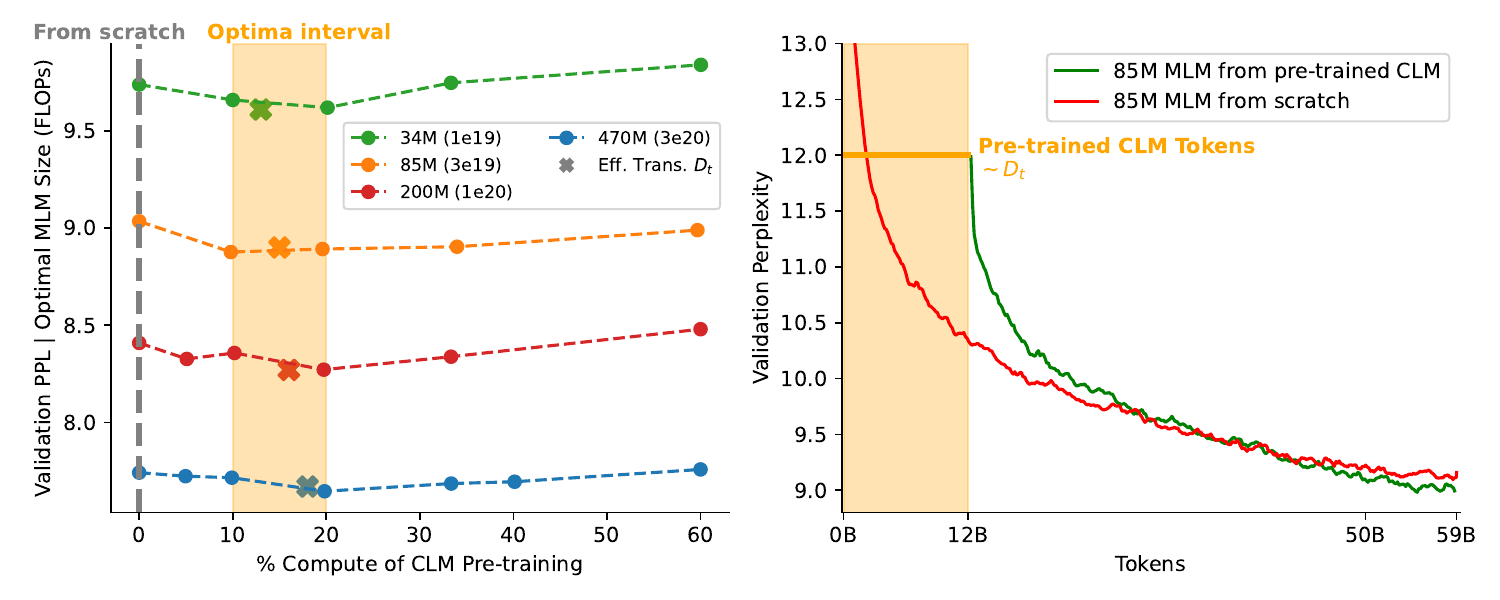}
	\caption{\textbf{Left}: Valid perplexity of \% compute allocated for the CLM pre-training. For instance, \% compute indicates first training on CLM and then the rest compute fine-tuning on MLM. The optimal CLM pre-training \% compute range with [10, 20]. And the fitted $ D_t / (D_t + D_f) $ drops in the optimal loss range. \textbf{Right}: Comparison of validation perplexity for models trained from scratch (red) and those fine-tuned from a pre-trained CLM (green), demonstrating that fine-tuning from a CLM reduces perplexity with similar or even fewer tokens. 
 } 
 \vspace{-3mm}

% The fine-tuning steps at least 10K? The winning tokens will diminish when model become larger, such as > 500M}
  \label{fig::opt_trans}
  %\vspace{-10pt}
\end{figure}

Although we observe that MLM benefits from transfer learning from CLM, the pre-training compute budget remains unaccounted for. We focus on two aspects: (1) the actual benefit CLM provides to MLM and its predictability, and (2) performance differences between MLM trained from pre-trained CLM (MLM-CLM) and MLM from scratch (MLM-S) under identical FLOP constraints.
We define Effectively Transferred Tokens $D_t$ as the \textit{additional data a model of the same size would need to train from scratch on MLM to achieve the same loss as a model pre-trained on CLM.} 
If the token number in the pre-trained CLM model exceeds $D_t$, then the computations for CLM pre-training was excessive. 
Knowing $D_t$ in advance would guide the allocation of tokens for CLM pre-training.

%Although we observe that MLM can benefit from transfer learning from CLM, the pre-training compute budget has not been accounted for. We are particularly interested in two aspects: (1) the actual benefit that CLM brings to MLM and whether this can be predicted, and (2) whether there is a difference in performance between MLM from pre-trained CLM~(MLM-CLM) and MLM from scratch~(MLM-S) under the same FLOP constraints.
%We propose the effectively transferred tokens $D_t$, \textit{specifically the additional data number that a model of identical size would require to train from scratch on MLM and achieve the same MLM loss as a model pre-trained on CLM.} 
%If the pre-trained CLM model's token number is significantly greater than $D_t$, then it is clear that computation was wasted on the CLM pre-training. 
%Thus, we propose a hypothesis: If we could know the $D_t$ in advance, it would inform us how to set CLM-pretraining tokens.

We compare MLM-S and MLM-CLM models ranging from 33M to 1.2B with FLOP counts from \(3 \times 10^{19}\) to \(1 \times 10^{21}\). By calculating the \textit{token distance} at the same loss level between these models, we establish our fitting target \(D_t\), collecting approximately 2800 sample points. Following similar methods in scaling transfer works~\cite{hernandez2021scaling,zhang2024scaling}, \(D_t\) is defined by a simple multiplicative scaling formula:
\begin{equation}
\label{eq::d_t}
D_t = k \times \frac{1}{D_f^\delta} \times \frac{1}{N^\gamma}; \quad k \approx 3.65 \times 10^5, \quad \delta \approx -0.137, \quad \gamma \approx -0.369 
\end{equation}
where \(D_f\) represents the tokens used for MLM-CLM, and $N$ is the number of parameters, with \(k\), \(\delta\), and \(\gamma\) as fitting coefficients. For instance, a 10$\times$ increase in \(D_f\) would roughly triple the model size and double \(D_t\).
We validate these findings by evaluating the compute ratio of CLM pre-training under four specified parameters and FLOPs, as shown in Figure~\ref{fig::opt_trans}~(left), finding that MLM-CLM generally outperforms MLM-S. Specifically, $D_t / (D_t + D_f)$ ranges from 10\% to 20\% of the compute budget for CLM pre-training. Figure~\ref{fig::opt_trans}~(right) schematically illustrates the learning curves of two 85M (3e19 FLOPs) models, with MLM-CLM achieving similar or better loss levels with equal or fewer tokens.

%when FLOPs <= 1e20.
%—or winning tokens—demonstrating an improvement over %starting from scratch with the same FLOPs count, approximately from 5B to 15B tokens in our experiments.
%\cxy{Emphasize the effective pre-training tokens. Weakening winning tokens, it may be diminish advance when training >= 1B models.}

\begin{findings}
Training MLM from scratch with 10× the compute requires approximately 7.7× the compute compared to MLM from CLM pre-training, implying that around 20\% of the compute budget should be allocated for CLM pre-training to get better MLM models transferred from CLM pre-training.
\end{findings}

\section{Experimental Validation}
Based on the scaling laws we observe, we estimate the model size and training tokens for current leading models by analyzing their FLOPs. In our configuration, the PROGEN2-xlarge model, with 6.4B parameters, is estimated to require training with 7.2B parameters and 265B tokens. Similarly, the ESM-2 model, with 3B parameters, should be trained with a model size of 10.7B parameters and 260B tokens. 
Additionally, we employed two 470M models to test the transfer scaling strategy, one trained from scratch~(470M scratch) and the other from CLM pre-training~(470M trans.).
The model's details are reported in Table~\ref{tab:model_comparison}.

\begin{table}[h]
\vspace{-10pt}
\centering
\small
\caption{\textbf{Model architecture details.} We compare popular models PROGEN2 and ESM-2 using similar FLOPs with our models estimated by proposed scaling law.} %showing model size, architectures, and training tokens.}
\begin{tabular}{@{}lcccccc@{}}
\toprule
Params & Objective & $N_{\text{head}}$ & Dim.  & $N_{\text{layer}}$ & Train. Tokens & FLOPs \\ 
\midrule
PROGEN2-xlarge~(6.4B) & CLM     & 16      & 4096  & 32  & 350B        & $1.34 \times 10^{22}$ \\
Our 7.2B          & CLM            & 32      & 4096 & 36  & 265B        & $1.14 \times 10^{22}$ \\
\hline
ESM-2~(3B)      & MLM            & 40      & 2560  & 36  & 1T          & $1.68 \times 10^{22}$ \\
Our 10.7B          & MLM            & 32      & 4352  & 47  & 260B        & $1.68 \times 10^{22}$ \\
\hline

470M scratch & MLM      &   16   & 1280  &  24 & 106B & $3.0 \times 10^{20}$ \\
%6.2B~(Section~\ref{sec::optimal_transfer})           & Joint      & 28     & 3584  & 24  & 46B + 182B & $8.4 \times 10^{21}$ \\
%470M Trans.1         & CLM + MLM      & 16     & 1280  & 24  & 35B + 106B & $ 4.0 \times 10^{20}$ \\
470M Trans.         & CLM + MLM      & 16     & 1280  & 24  & 21B + 85B & $ 3.0 \times 10^{20}$ \\

%6.16B           & CLM + MLM      & 28     & 3584  & 24  & 230B + 420B & $1.68 \times 10^{22}$ \\
%6.16B~\ref{section:mask-strategy}           & CLM + MLM      & 28     & 3584  & 24  & 30B + 420B & $1.68 \times 10^{22}$ \\

%ESM-2~(650M)  & MLM      & 20     & 1280  & 33  & 1T & $3.9 \times 10^{21}$ \\
%6.2B~(Section~\ref{sec::optimal_transfer})           & Joint      & 28     & 3584  & 24  & 46B + 182B & $8.4 \times 10^{21}$ \\
%Our 3.4B          & CLM + MLM      & 22     & 2816  & 22  & 38B + 155B & $ 3.9 \times 10^{21}$ \\

\bottomrule
\end{tabular}
\label{tab:model_comparison}
\vspace{-10pt}
\end{table}

%We use a similar FLOP count with PROGEN2-xlarge~(6.4B) and ESM-2 (3B). Our 7.2B and 10.7B model size and training tokens allocation are predicted by power-lar in Equation~\ref{eq:c_law} and 3.4B MLM is decided by adopting the transfer scaling strategy where first train 38B tokens and then train 155B tokens. 
%\cxy{rewrite, emphasize same flops, what is the optimal model.}
%our evaluation 主要集中在LM中主流的benchamarks上，例如Contact prediction, Fold Prediction.
%为了尽可能减少下游评估pipline的带来的不确定性，仅使用，unsupervised contact prediction，以及linear probing来观察我们的protein engineering tasks. 这样更加凸显预训练模型的基础能力，
%对于CLM等生成式的模型，我们主要对比不同模型的生成样本和真实蛋白序列的比较情况。 
%根据ESM-2, PROGEN2等模型所消耗的FLOPs=6ND，我们基于scaling law 使用这些FLOPs推算出了3个模型通过计算我们的power-law, 分别是和ESM-2 3B同等算力的10.7B MLM 模型，和PROGEN2-xlarge类似算力的7.2B CLM模型， 以及和ESM-2 （650M)对比的3.4B CLM + MLM的模型。

%Our scaling law experiments are under 1 epoch training setting.
%following the data-constrained scaling law and our repeat experiments~(section 1), we control the training tokens under the near 1 epoch, it will not  excess the 25\% of the dataset size. 

%super linear scale

%\subsection{Protein Generation Comparison with CLM}
\subsection{Protein Generation Comparison: 7.2B CLM vs. 6.4B PROGEN-xlarge}

\begin{figure}
	\centering
\includegraphics[width=0.99\textwidth]{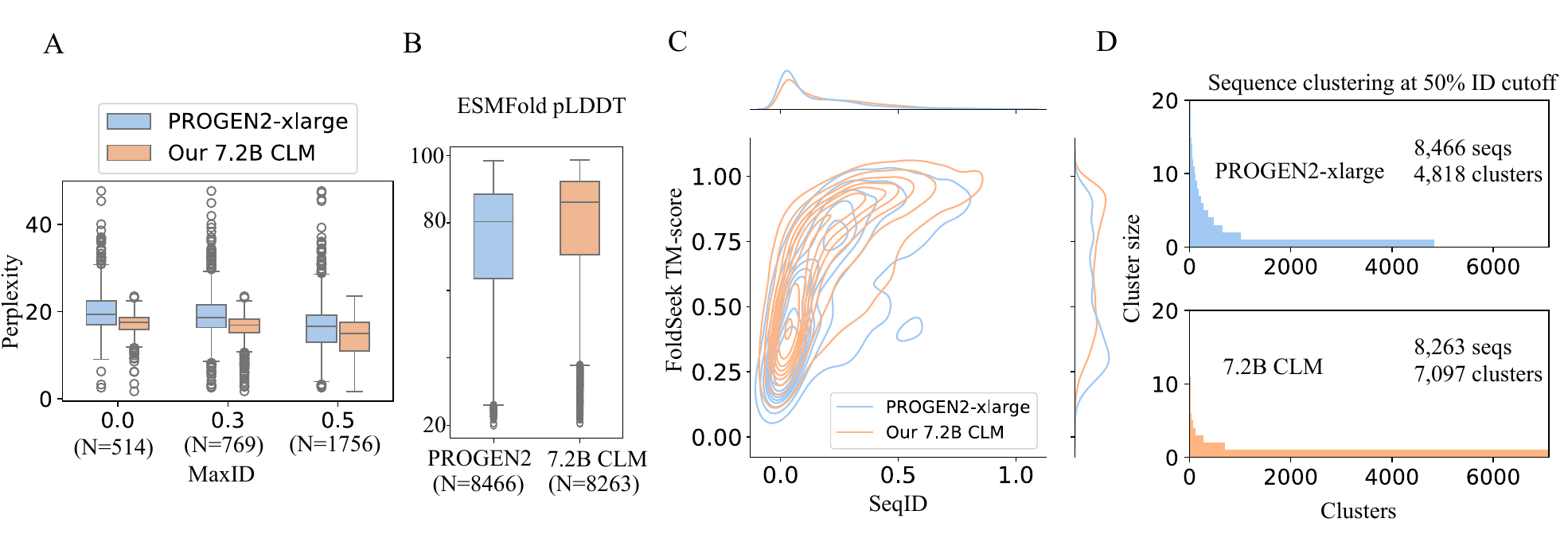}
\vspace{-3mm}
\caption{\textbf{Comparative Analysis of CLM Models.}
\textbf{A.} Perplexity analysis for PROGEN2-xlarge and our 7.2B CLM shows lower values for our model across various MaxID levels, suggesting better sequence handling.
\textbf{B.} Box plots of pLDDT scores for protein structures by PROGEN2-xlarge and our 7.2B CLM.
\textbf{C.} Contour and line plots show our 7.2B CLM sequences mimic natural sequences more closely than PROGEN2-xlarge, assessed using Foldseek with the PDB database.
\textbf{D.} Clustering at 50\% sequence identity reveals our 7.2B CLM generates more clusters, indicating higher diversity.}

% The fine-tuning steps at least 10K? The winning tokens will diminish when model become larger, such as > 500M}
  \label{fig::clm_task}
  \vspace{-20pt}
\end{figure}

We first evaluate the perplexity on OOD data and then compare the protein generation capabilities of the 7.2B CLM and PROGEN2-xlarge models. Each model generated 2,000 sequences for each parameter combination of top-$p$ $\{0.5, 0.7, 0.9, 1.0\}$ and temperature $t$ $\{0.2, 0.4, 0.6, 0.8, 1.0\}$, totaling 40,000 sequences per model. Sequences with a perplexity greater than 10 and duplicates were removed, leaving 8,263 and 8,466 sequences for the 7.2B CLM and PROGEN-xlarge, respectively. 
We used four metrics to assess the quality of the models and the generated sequences (See Appendix~\ref{app:gen_eval} for details).

%We first evaluate the perplexity on out-of-distribution data. And we also compared the protein generation capabilities of the 7.2B CLM and PROGEN2-xlarge models. Each model generated 2,000 protein sequences for each combination of the parameters top-$p$ $\in \{0.5, 0.7, 0.9, 1.0\}$ and Temperature $t$ $\in \{0.2, 0.4, 0.6, 0.8, 1.0\}$, totaling 40,000 sequences per model. The sequences with perplexity (PPL) greater than 10 and those contain duplicates are removed. After this filtering process, 7.2B CLM and PROGEN-xlarge retained 8,263 and 8,466 sequences, respectively. 
%We adopt four different metrics to measure the quality of the models and generated sequences:

%\panli{'PPL is not evaluated on generated data'}

\vspace{-5pt}
\vpara{OOD Dataset PPL Analysis}
We randomly sampled 5,000 sequences from UniProt released after 2023-01-01 and aligned them to our and PROGEN2's training data (Uniref90 and BFD) using HHblits~\cite{remmert2012hhblits} or Jackhmmer~\cite{ebi_jackhmmer}. Sequences below a maximum identity cutoff were used to assess the models' PPL, as shown in Figure~\ref{fig::clm_task}A. Our 7.2B CLM exhibited lower PPL on three subsets.

\vspace{-5pt}
\vpara{pLDDT scores from ESMFold}
Atomic structures of 8,263 and 8,466 generated sequences were predicted using ESMFold, and compared based on pLDDT scores, displayed in Figure~\ref{fig::clm_task}B. The 7.2B model's average pLDDT score was 78.69, higher than PROGEN2-xlarge's 74.33.

\vspace{-5pt}
\vpara{Natural Sequences Comparisons with Foldseek}
Using Foldseek~\cite{van2022foldseek}, we searched the PDB database for sequences similar to those generated by our 7.2B CLM model, which showed better mimicry of natural sequence properties with higher average TM-scores (0.655 vs 0.522) and SeqID (0.194 vs 0.165), as shown in Figure~\ref{fig::clm_task}C.

\vspace{-5pt}
\vpara{Diversity Analysis}
Generated sequences were clustered using MMseqs2~\cite{steinegger2017mmseqs2} with a 50\% similarity cutoff. The 7.2B CLM model resulted in higher diversity with 7,097 clusters compared to 4,818 clusters for PROGEN2-xlarge, detailed in Figure~\ref{fig::clm_task}D.

\vspace{-10pt}
\subsection{Protein understanding tasks: 10.7B MLM vs. 3B ESM2}
%~\cite{rao2020transformer}
\vspace{-5pt}
We evaluate different task types from the protein benchmark~\cite{chen2024xtrimopglm}: Contact prediction as binary classification at the amino acid pair level; fold classification into 1195 classes at the sequence level; and fluorescence as regression tasks. Following ~\cite{chen2024xtrimopglm}, we add a Multi-Layer Perceptron~(MLP) head to each pre-trained model and apply Low-Rank Adaptation (LoRA)~\cite{hu2021lora}~(r=8, $\alpha$=16) for fine-tuning 
(see Appendix~\ref{app:down} for convergence details).

\begin{table}[h]
\vspace{-10pt}
\centering
\small

\caption{Tasks performance of MLM Model on the test dataset with LoRA fine-tuning.}
\begin{tabular}{@{}lcccc@{}}
\toprule
Models & Contact Pred.~(P@L/5) & Fold Class.~(1195 class.) & Fluor.~(reg.) \\% &  Stab.~(reg.)  \\ 
\midrule
ESM-2~(3B)  & 0.91    & 0.69           &   0.65   \\% &  0.81   \\
Our 10.7B   & 0.91       & \textbf{0.72}            &  \textbf{0.69} \\%& \textbf{0.82}   \\
\midrule
470M scratch   & 0.78       & 0.65            &  0.67 \\%&0.74   \\
%470M Trans.-1  & 0.81       &  0.68            &  0.69 & 0.82   \\
470M trans.  & \textbf{0.80}       & \textbf{0.66}            &  0.67 \\% & 0.74   \\

%\hline
%6.4B (PROGEN2-xlarge) & CLM     & 16      & 4096  & 32  & 350B        & $1.34 \times 10^{22}$ \\
%7.2B~(Table~\ref{tab:coeff})           & CLM            & 32      & 4096 & 36  & 265B        & $1.20 \times 10^{22}$ \\
%\hline
%6.16B           & CLM + MLM      & 28     & 3584  & 24  & 230B + 420B & $1.68 \times 10^{22}$ \\
%6.16B~\ref{section:mask-strategy}           & CLM + MLM      & 28     & 3584  & 24  & 30B + 420B & $1.68 \times 10^{22}$ \\

%ESM-2~(650M) & 90.1  & 64.6     &  -    & 79.8   \\
%6.2B~(Section~\ref{sec::optimal_transfer})           & Joint      & 28     & 3584  & 24  & 46B + 182B & $8.4 \times 10^{21}$ \\
%Our 3.4B   &        &  \textbf{66.4}     & \textbf{67.1}     & -    \\

\bottomrule
\end{tabular}
\label{tab:ds_perform}
\vspace{-10pt}
\end{table}

The results, shown in Table~\ref{tab:ds_perform} and \ref{tb:arti_app}, demonstrate that our 10.7B model outperforms ESM-3B on 7 out of 8 tasks. This confirms the rationale behind the observed scaling law and addresses concerns about the scope and rigor of our evaluation tasks.
Additionally, the 470M model transferred from CLM pre-training continues to perform effectively in this task, showing the efficacy of the observed transfer scaling law.

\section{Discussion and Limitations}
\label{app:dis}

\vpara{Data Repeat Scaling Law.}
Our scaling law is learned within a single epoch setting. It is well known that MLM exhibits higher sample efficiency than CLM due to the dynamic masking strategies across epochs. However, this advantage diminishes when training is limited to only one epoch. 
% This also suggests that for MLM training, a small amount of repetition can be considered as new data, without detriment to the performance.
We present an empirical study by comparing a 2.8B model trained on 1T tokens (approximately five epochs) against a 10.7B model trained on 265B tokens (roughly 1.4 epochs). Despite the models utilizing the same amount of FLOPs, the two models achieve similar capability in terms of OOD PPL (10.33 vs 10.21).
% Despite this, the impact of training MLM for several epochs repeatedly is not significant in terms of loss. 
% This insight suggests that repeating several rounds under MLM training has a minimal impact on reducing loss, and our scaling law does not necessarily need to be confined within 200 billion tokens.
While the smaller models are more user-friendly during inference and fine-tuning. Therefore, we also suggest an alternative approach that adjusts the optimal training token count and model size within the data repeat scaling law.
% scaling law framework when scaling MLM. We will further investigate repeat scaling laws as designated in future work~\cite{muennighoff2024scaling}.

%repetitions with MLM may have better performance than one-epoch training, Specifically on a specific task

\vpara{Multi-modality Scaling.}
% We observe that the scaling laws for CLM, also known as autoregressive models, exhibit similarities to those in natural languages or the code modality in the context of protein sequences, closely aligning with findings by Chinchilla~\cite{hoffmann2022training}. 
The multi-modal auto-regressive work~\cite{henighan2020scaling} suggests the existence of a nearly universal scaling law across various modalities, including images, videos, math, code, and languages. Our results appear in this trend as well, such as, the scaling laws for CLM exhibit similarities to those in natural languages. 
The same situation may apply to other modalities of biological data, such as RNA and DNA~\cite{nguyen2024sequence}. 

\vpara{Various Pre-train Datasets and Strategies.}
Our datasets cover a substantial portion of the protein universe, yet they might not be entirely representative. Combining BFD~\cite{bfd_database}, Uniref~\cite{suzek2015uniref}, MetaClust~\cite{levy2020metaeuk}, and IMG/JGI~\cite{markowitz2006integrated} with 90\% clustering results in at least 600 billion unique tokens. However, variations in datasets could affect the power-law behavior.
Future work could explore applying our findings to different model architectures. There is ongoing research on scaling LLMs for long sequences~\cite{beltagy2020longformer, child2019generating, choromanski2020rethinking,dao2022flashattention,jacobs2023deepspeed,liu2023ring,zaheer2020big}, and MSA augmentation could significantly improve protein representation regarding contacts and structure. Investigating scaling laws in this context could be a promising direction for future research.

\section{Conclusion}
In this work, we are the first to establish a practical pathway for researchers to develop faithful and powerful protein language models optimized by both CLM and MLM objective in an end-to-end manner. This includes everything from pre-training dataset construction, expanded metagenomic databases such as ColabFoldDB, emphasizing the critical importance of data quality and quantity for scaling language models, to optimal parameter and dataset allocation along with the potential loss prediction, as well as knowledge transfer from other pre-training objectives. Our work holds significant potential for the application of large language models across various scientific domains.

\vpara{Acknowledgments.}
This work has been supported by the National Key R\&D Program of China 2021ZD0113304, NSFC for Distinguished Young Scholar 62425601, New Cornerstone Science Foundation through the XPLORER PRIZE.

\bibliographystyle{plain}
\bibliography{ScalingPLMs}  %%% Uncomment this line and comment out the ``thebibliography'' section below to use the external .bib file (using bibtex) .

\clearpage

\newpage
\appendix
\renewcommand{\thefigure}{A\arabic{figure}}
\renewcommand{\thetable}{A\arabic{table}}
\part{Appendix}
\parttoc

%ROPE, Flash Attention
%we rely on existing work and provided experimental heuristics to determine the other necessary hyperparameters

\begin{table*} % 使用 [H] 选项强制在当前位置放置
    \newcolumntype{?}{!{\vrule width 1pt}}
    \newcolumntype{C}{>{\centering\arraybackslash}p{5em}}
    \caption{
        \label{tb:arti_app} Tasks performance of MLM Model on the 5 test dataset, 
        Fitness Prediction (Fit P.) as a regression task measured by Spearman coefficient at sequence level,
        Localization (Loc.) as 10 sub-cellular classification task at sequence level,
        Metal Ion Binding (MIB) as a binary classification task at sequence level,
        Solubility (Sol.) as a binary classification task at sequence level,
        Stability (Sta.) as a regression task measured by Spearman coefficient at sequence level, 
        with LoRA fine-tuning.
    }
    \footnotesize
    \centering 
    \renewcommand\arraystretch{1.0}
    \begin{tabular}{@{~}l?@{~}*{1}{CCCCC}@{~}}
        \toprule
        \textbf{Model} & \textbf{Fit P. (SP)} & \textbf{Loc. (ACC)} & \textbf{MIB (ACC)}& \textbf{Sol. (ACC)}& \textbf{Sta. (SP)}\\
        \midrule
        ESM2 (3b)
        & 0.94&\textbf{0.81}&0.82&0.74&0.82\\
        Our 10.7B
        & \textbf{0.96} & 0.79 & \textbf{0.83} & \textbf{0.79} & \textbf{0.83} \\
        \bottomrule
    \end{tabular}
    % \vspace{-20pt}
\end{table*}

\section{Related Work}

\label{Section:Related Work}

\vpara{Protein Language Model}
Since the advent of AlphaFold2~\cite{alphafold}, the masked language model (MLM) has been integrated as a subtask within the Evoformer architecture. In this context, an assumption is that large language models can be considered as a lossless compression method~\cite{deletang2023language}. This was followed by a series of language modeling efforts~\cite{ferruz2022protgpt2, brandes2022proteinbert, heinzinger2023prostt5, elnaggar2021prottrans, elnaggar2023ankh}, which aimed to conduct pre-training on single-sequence proteins using larger datasets and model scales. These efforts sought to harness the scale of the models to learn complex co-evolutionary information, although detailed investigations on how to optimally scale these models remain scarce. Our work primarily focuses on these finer aspects, aiming to fill this gap in the research.

%Since the introduction of AlphaFold2~\cite{alphfdol2}, masked language models (MLM) have been utilized as a subtask within the Evoformer architecture. It is posited that large language models can act as a lossless compression method for protein data~\cite{compression}. This has prompted a series of language modeling initiatives~\cite{ESM,probert,prot5,xtrimopglm,omegaplm,protrans,ankh}, which have focused on pre-training on single-sequence proteins with larger datasets and model scales. These efforts aim to leverage the scale of the models to learn complex co-evolutionary information. However, detailed investigations on how to optimally scale these models are still lacking. Our work primarily focuses on these finer aspects, aiming to fill this gap in the research.

%To date, no pre-trained model has surpassed AlphaFold2 in terms of structural prediction accuracy.
%Despite these challenges, from the perspective of large language models as lossless compressors, with advancements in computational power and the evolution of language models, incorporating homologous sequences into long-sequence fine-tuning from current language model could be feasible.

\vpara{Training objectives}
In natural language processing (NLP), masked language models (MLM) are rarely adopted due to the self-explanatory nature of natural language, which inherently prompts the meta-knowledge of tasks and generates task targets through CLM (Conditional Language Modeling) training models.
However, a unified language modeling objective for Protein Language Models has yet to be fully consented. Those based on causal language modeling (CLM) have been primarily explored for protein design. Benchmarks in protein design using MLM~\cite{wang2019bert} have also shown promising results for generation~\cite{notin2024proteingym}, exhibiting variable performance when compared to CLM~\cite{zheng2023structure, verkuil2022language}. Additionally, the potential of the in-filling task objective remains largely unexplored~\cite{bavarian2022efficient,tay2022ul2,du2021glm}.
Our research aims to thoroughly discern the scaling behavior of the two most common optimization objectives in this domain.

%We hope to see a future where a unified self-supervised optimization goal and model architecture can define the Foundation Model in the biological domain.

\vpara{Scaling Laws}
To our knowledge, the concept of scaling laws of language model is first introduced by OpenAI~\cite{kaplan2020scaling}. Subsequently, numerous variants and modifications~\cite{hoffmann2022training} have been developed around this theme. Recently, an array of new scaling laws has emerged. These include scaling laws related to learning rates and batch sizes~\cite{bi2024deepseek}, data-constrained scaling laws~\cite{muennighoff2024scaling}, scaling laws for downstream tasks and Transfer~\cite{zhang2024scaling,hernandez2021scaling}, as well as scaling laws within the Mixture of Experts (MoE) framework~\cite{clark2022unified}, and those concerning long sequences and positional encoding~\cite{liu2023scaling}.  While these laws are primarily derived using auto-regressive models in resource-rich domains, their application in the biological data sector is less common. Our work seeks to address this gap. Furthermore, scaling laws for Masked Language Models (MLM) are notably scarce. Given that MLMs are currently one of the most effective training methods for biological data, our research on MLMs could also be extended to other non-text domains.

\section{UR50/S Repeat Experiments}
\label{app:repeat}

\begin{figure}[ht]
	\centering
\includegraphics[width=0.99\textwidth]{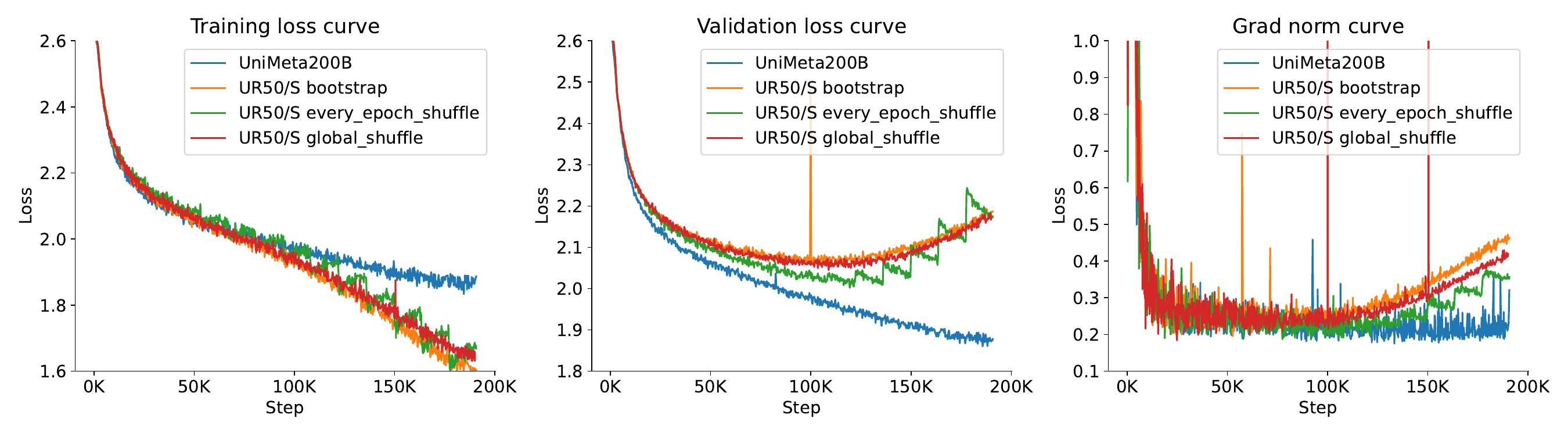}
	\caption{\textbf{Learning curve for UR50/S dataset repetition methods}. Our 194B tokens dataset~(UniMeta200B) shown in blue, serves as the reference with an approximate single epoch run. The bootstrapping method, depicted in orange, processes 200 billion tokens with replacement, indicating a tendency towards zero unsampled tokens by the fifth epoch. The every-epoch shuffle method, in green, ensures all tokens are used per epoch, forming a stair-step pattern in training loss. Lastly, the global shuffle method, in red, loosely uses all tokens each epoch but ensures the strict number of epoch passes for every token. The rightmost plot of gradient norms shows an uptick for curves corresponding to overfitting, signifying a lack of further optimization, with steep or erratic gradients indicated by the ascending gradient norms.} 
% The fine-tuning steps at least 10K? The winning tokens will diminish when model become larger, such as > 500M}
  \label{fig::overfit_ur50}
\end{figure}

We employed three different methods to repeat training on the UR50/S dataset, all of which ultimately led to overfitting. The reference for these experiments is shown by the blue curve in Figure~\ref{fig::overfit_ur50}, which represents UniMeta's loss for approximately one epoch.

Firstly, using bootstrapping, we processed 200 billion tokens from UR50/S with replacement. In each epoch, 65\% of the dataset was randomly selected, leading to a diminished proportion of unsampled tokens by the fifth epoch, as depicted by the orange curve.

Secondly, we shuffled the unique data for each epoch to ensure that all UR50/S tokens were used per epoch, resulting in a stair-step pattern~\cite{fastai_learning_jumps} in the training loss, illustrated by the green curve. It has simply memorized the dataset but isn’t improving at generalizing. Over-confident predictions of the first batch of the next epoch lead to a big step update, and then the model is not adapted to the next batches, resulting in no longer a decrease in loss.

Lastly, we shuffled the entire training dataset less stringently, which did not strictly ensure that all UR50/S tokens were used every epoch, but guaranteed that each token was used an equal number of times over the entire training period. We term it global shuffle, this approach is shown by the red curve.

From the gradient norm curve shown in Figure~\ref{fig::overfit_ur50}~(right), we observe an uptick in gradient norm for the overfitting curves, indicating that the model is no longer optimizing effectively. In machine learning, such an increase in gradient norm typically suggests that the model is encountering areas of the parameter space where gradients are steeper or more erratic, often occurring when the model starts to memorize the training data rather than generalize from it, approaching a saturated network~\cite{merrill2020effects}. This behavior can result from overly complex models, too many training epochs without sufficient regularization, or training on non-representative data.

%We conduct three repeat methods on Uniref50 training, overall it eventually ovefit.
%\item Boostrapping, we run 200B tokens on unique Uniref50, ex`xvery token repeated uneven times. Specifically, each epoch randomly takes 65\% of the unique dataset. At 5 epoch, the unsampled tokens proportion will tend to zero.

%\item Unique data is shuffled for each poch, it strictly ensure passing  uniref50 tokens for every epoch. it will develop a stairs-like training loss.

%\item shuffle on the whole training samples, it unstrictly ensure passing all uniref50 tokens for every epoch but eventually every token will be training strict epoch times. 

\section{Choice of Masking Ratio}
\label{app:mask_ratios}
\begin{figure}[!thbp]
  \centering
  \subfloat{\includegraphics[width=0.5\textwidth]{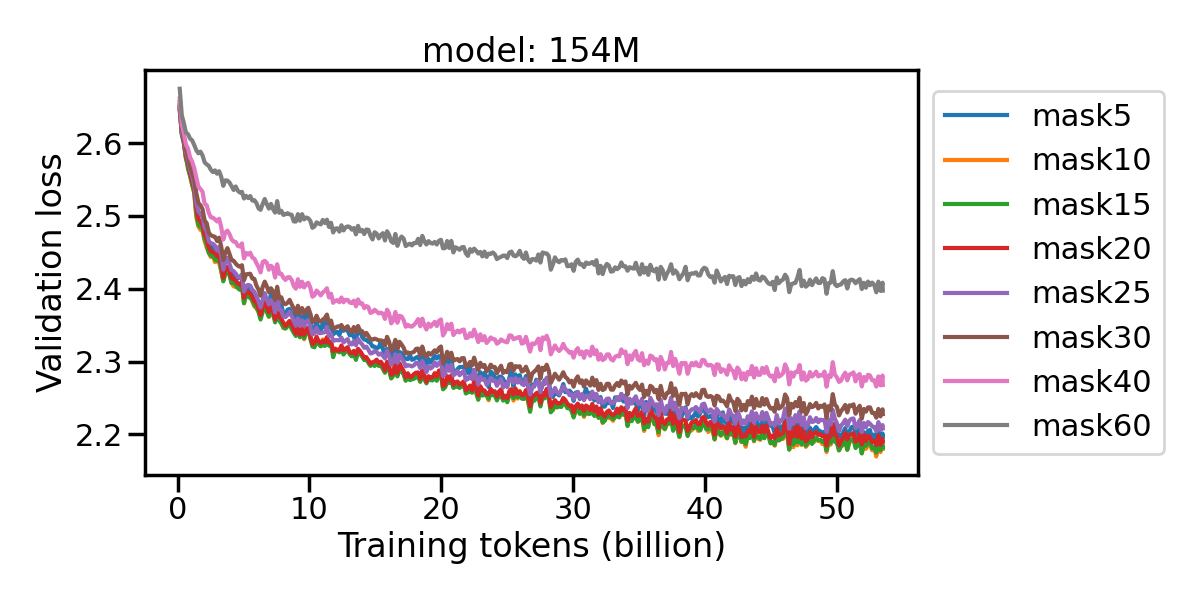}}
  \hfill
  \subfloat{\includegraphics[width=0.5\textwidth]{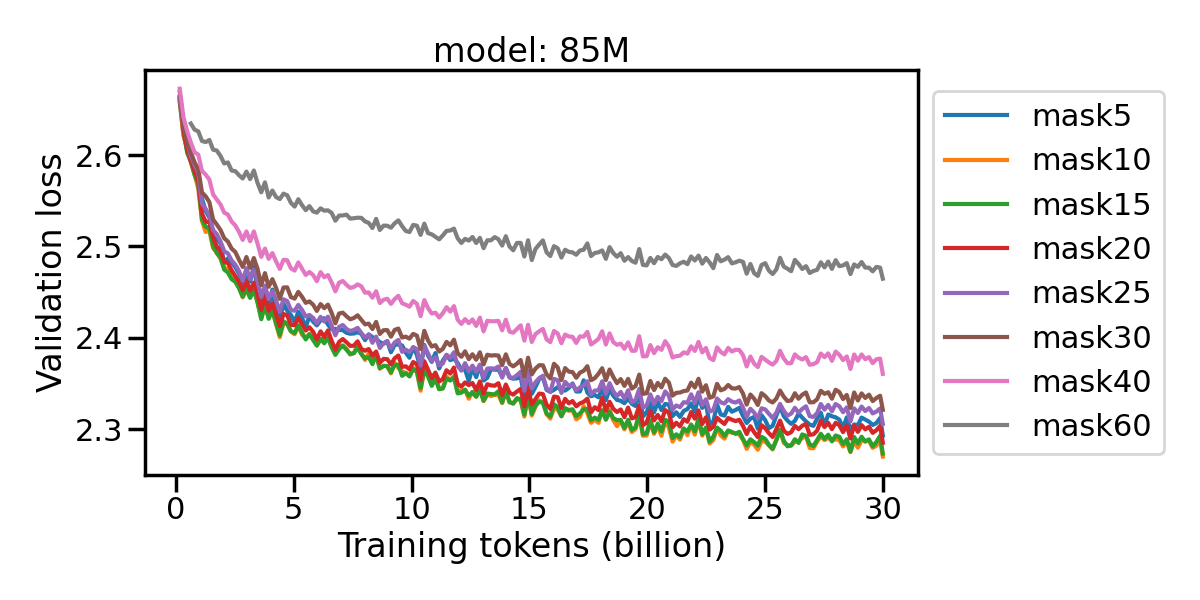}}
  \caption{\textbf{Validation loss of different masking ratios}. Two models (154M and 85M) are trained from 5\% to 60\% masking intervals.}
  \label{fig:mask-ratio-val-loss}
\end{figure}

\begin{figure}[!h]%{width=0.9\textwidth}
    \centering
%    \vspace{-40pt} 
    \includegraphics[width=0.8\textwidth]{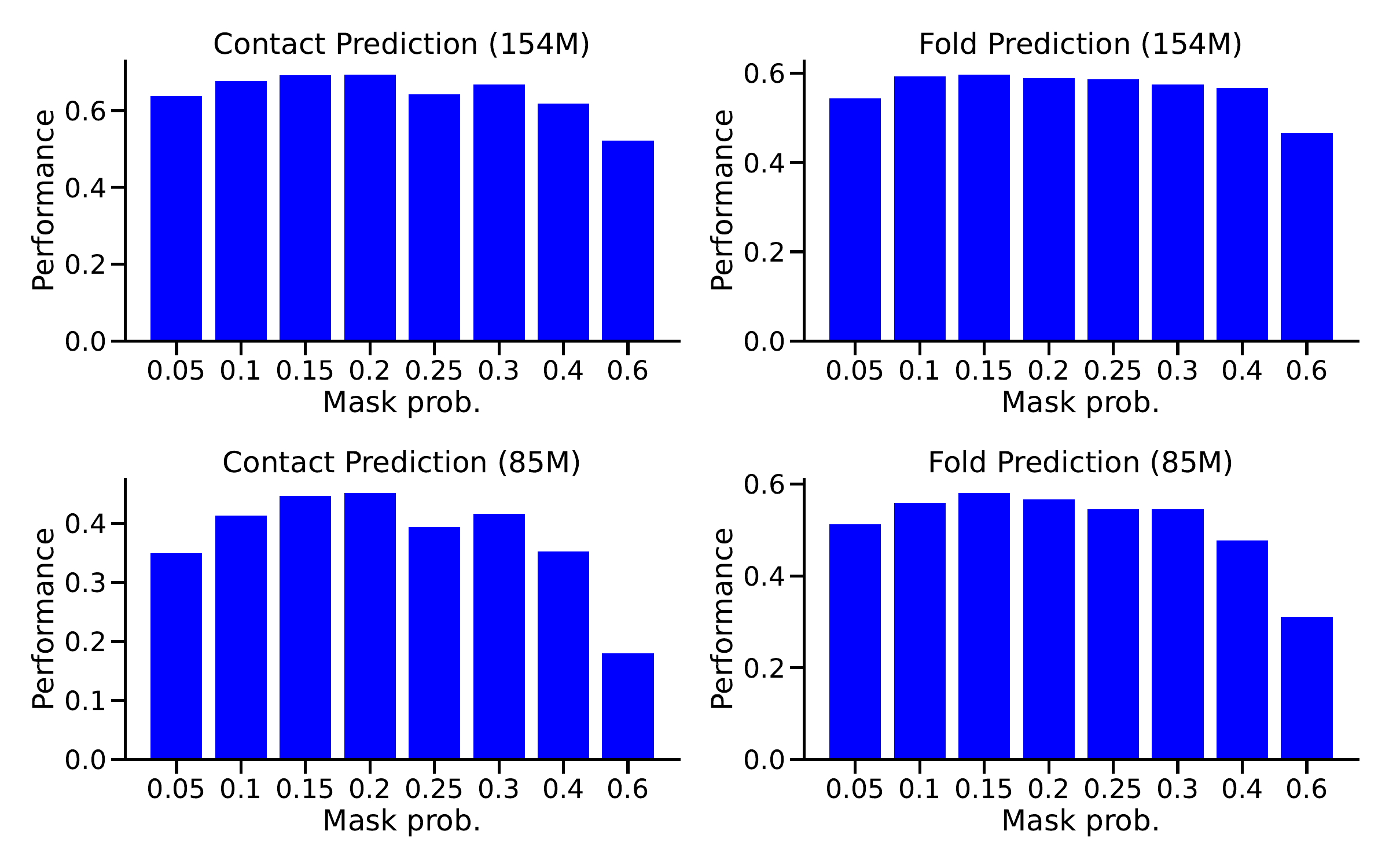}
    \caption{\textbf{Abalation of different masking ratios}. Two models (154M and 85M) are trained from 5\% to 60\% masking intervals, and evaluated on contact map and fold classification downstream tasks.}
    \label{fig:mask-ratio}
   % \vspace{-40pt} 
\end{figure}

In the original BERT work~\cite{BERT}, the absence of masked tokens in downstream tasks presented a mismatch with the pre-training data distribution. The authors investigated various masking ratios and concluded that a 15\% masking rate was most beneficial for downstream tasks. This was implemented alongside an 80-10-10 strategy: 80\% of the tokens were replaced with a mask, 10\% were randomly substituted, and the remaining 10\% were left unchanged. 

However, given the significant differences between protein sequences and natural language processing data, we employed two models, sized at 85M and 154M, to explore a range of masking ratios from 5\% to 60\%~(see Figure~\ref{fig:mask-ratio-val-loss}). The best masking ratios for validation loss drop ranged from 10\% to 20\%; ratios too small (5\%) or too large (greater than 25\%) degraded the performance.

We further used pre-trained eight different models to perform full fine-tuning on downstream tasks such as Contact Prediction and Fold Classification in Figure~\ref{fig:mask-ratio}. Results from the test datasets revealed that, similar to NLP, the optimal performance was achieved within a 10\%-20\% masking range. Specifically, a 20\% masking ratio slightly outperformed 15\% in Contact Prediction, while the 15\% ratio yielded the best results in Fold Prediction. Consequently, for our Masked Language Model (MLM), we decided to adhere to the 15\% masking ratio with the 80-10-10 strategy for training all our models.

%\begin{table}[h]
%\centering
%\caption{$D_t$ is the CLM pre-training tokens, which predicted by equation~\ref{eq::d_t}, $D_f \approx  \left[\text{FLOPs}/(6 * \text{params}) - D_t\right]$}
%\begin{tabular}{@{}lccccc@{}}
%\toprule
%Params &  $ D_t$ &$D_f$ & FLOPs & PPL & MLM f. s. PPL   \\ 
%\midrule
%34M    &  $6.4B$  & $42.6B$ &1e19 & $9.76$  & $9.86$\\
%85M    &  $9.2$ & $49.8B$ & 3e19 & 8.97 & 9.17 \\
%200     &  $13.2B$ & $70.1B$ &1e20 & 8.27 & 8.41 \\
%470M     &  $18.7B$ & $87.6B$ &3e20 & 7.67 & 7.86 \\
%1.2B   &  $27.5B$  & $ 111.3B$  & 1e21 & &\\

%\bottomrule
%\end{tabular}

%\label{tab:coeff}
%\end{table}

%we use the $N^{\text{MLM}}(C), C \in {1e19,3e19,1e20,3e20,1e21}$, vary the $pre-training FLOPs$

\section{MLM/CLM for Protein Contact Prediction}
\label{app:mlm_vs_clm}

\begin{figure}[!h]%{width=0.9\textwidth}
    \centering
%    \vspace{-40pt} 
    \includegraphics[width=0.8\textwidth]{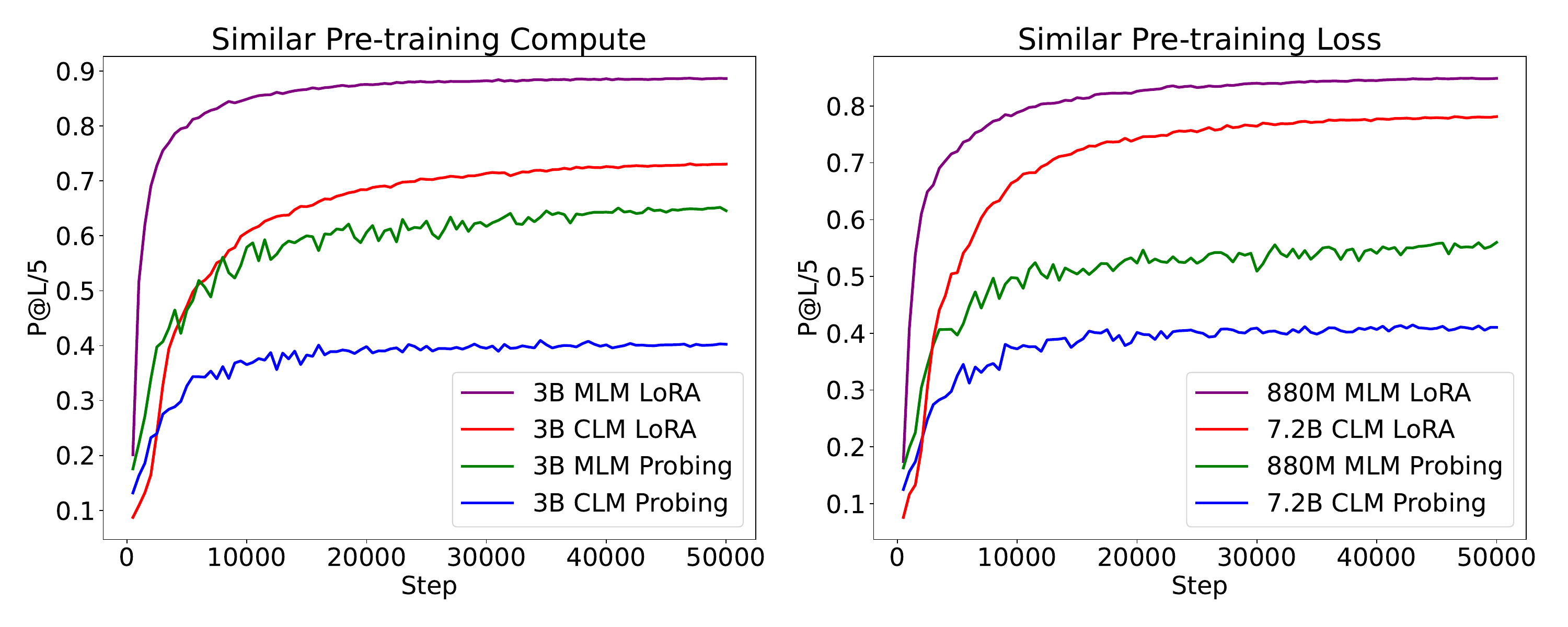}
    \caption{\textbf{Contact Prediction on MLM and CLM models.} Two 3B models (CLM and MLM) were trained using identical computational resources, represented by the probing and LoRA fine-tuning methods. On the right, performance of a 7.2B CLM model is compared with an 880M MLM model under similar pre-training loss conditions. These models exhibit differing rates of convergence, highlighting the impact of uni-directional and bi-directional model architectures on learning dynamics.
}
    \label{fig:mlm_vs_clm}
   % \vspace{-40pt} 
\end{figure}

We compared the effectiveness of CLM in the downstream task of contact prediction, using two different setups~(Figure~\ref{fig:mlm_vs_clm}). In the first setup, two 3B models were trained under identical computational resources on 200 billion tokens, $3.4 \times 10^{21} FLOPs$. Their performance was evaluated through two training approaches: Probing (freezing the pre-trained model) and LoRA fine-tuning, with an added MLP head for comparison.

In the second setup, we compared the effects of MLM and CLM under similar loss conditions. Here, a 7.2B CLM model and an 880M MLM model were selected, both achieving a loss of 1.98 on our validation set. Despite the MLM model having a simpler loss calculation, involving a 15\% mask rather than a one-by-one mask—which would result in a higher loss—the MLM significantly outperformed the CLM. Importantly, the CLM model's computational power was an order of magnitude greater than the MLM model~($1.68 \times 10^{22}$ vs $1.0 \times 10^{21}$ FLOPs). This suggests that despite the lower loss achievable by the CLM model compared to MLM with a one-by-one mask, the unidirectional limitations of CLM do not translate into better downstream task performance.

\section{Pre-training Dataset Quality}
\label{app:data_quality}

Compared to Uniref90, ColabFoldDB offers a higher diversity and larger numbers of protein sequences, though with generally shorter sequence lengths, likely suggesting potentially lower data quality. To evaluate the efficacy of our expanded dataset, ColabFoldDB, we initially trained two 85M models separately on Uniref90 and ColabFoldDB. Uniref90 in our dataset comprises two subsets: Uniref50/S and the incremental dataset over Uniref50/S, termed Uniref90/50. Similarly, ColabFoldDB consists of representative and member data. We controlled the sampling proportion to ensure uniform sampling across both datasets, with results reported in Table~\ref{tab:data_quality}. Both models were then trained using identical configurations on a 50B scale.

From the perspective of validation loss in pre-training, the higher loss on ColabFoldDB might be attributed to its lower diversity and shorter sequence lengths compared to Uniref90. However, the performance on downstream tasks, such as contact prediction and fold classification, shows negligible differences between models trained solely on ColabFoldDB and those trained on Uniref90, as illustrated in Figure~\ref{app:data_quality}. This confirms that ColabFoldDB is an effective expansion of Uniref90 that maintains sample efficiency.

\begin{table}[h] 
\centering
\small
\caption{\textbf{Compared two dataset characteristics.} Protein sequence count, token number, and sampling proportions for Uniref50/S, Uniref90/50, and ColabFoldDB representative and member data.}
% \begin{small}
% \begin{sc}
\begin{tabular}{@{}lcccc@{}}
\toprule
 Datasets & Prot. Seq.  &  Tokens~(AAs) & Sampling Prop. \\
 \midrule
 Uniref50/S & 54M & 15.2B &   28.67\%  \\
 Uniref90/50 & 102M &  37.8B &  71.33\%  \\
  \midrule
 ColabFoldDB$_c$ & 208M & 37.7B & 26.75\% \\
 ColabFoldDB$_m$ & 575M & 103B &  73.52\%   \\
% Total & 939M & 194B &  -  \\
\bottomrule
\end{tabular}
\label{tab:data_quality}
\end{table}

%ur50/S 53578119 15B
%ur90/50 102321303 38B
%colab_c 207846004 38B
%colab_m  575713783 100B

\begin{figure}[!ht]%{width=0.9\textwidth}
    \centering
    \includegraphics[width=1.0\textwidth]{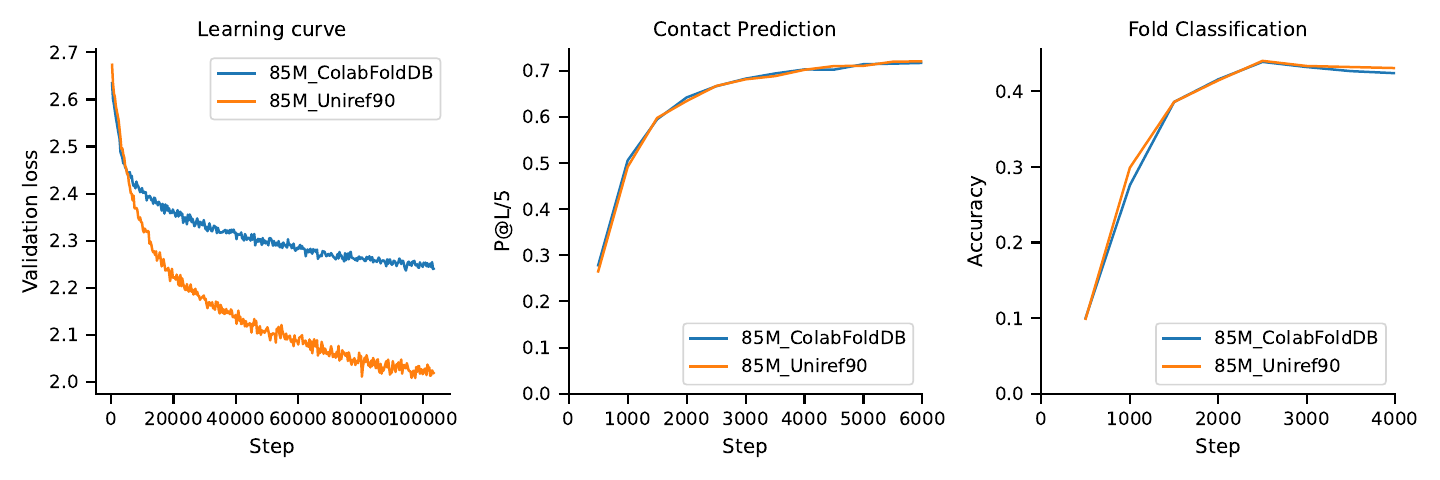}
    \caption{~\textbf{Data quality check.} Comparison of learning dynamics and downstream task performance for two 85M models trained on ColabFoldDB and Uniref90. Left: Validation loss curves demonstrating initial training differences. Middle: Contact prediction performance showing the response to testing on similar tasks. Right: Fold classification accuracy, comparing model responses to structural prediction tasks. Despite initial differences in loss, both datasets yield comparable performance in downstream applications.
}
    \label{fig:data_quality}
   % \vspace{-40pt} 
\end{figure}

\section{The Evaluation of Protein Generation.}
\label{app:gen_eval}
We explain more details about Protein Generation Comparison as follows:

\ipara{OOD Dataset PPL Analysis.} PPL represents the probability of the test sequences in the model distribution. The lower the PPL, the closer the distribution of the model and the test set is. In order to test the generalization ability of the model on new data, we use different sequence identity (0.0, 0.3, 0.5) as thresholds to select the test set.

\ipara{pLDDT Scores from ESMFold.} Predicted Local Distance Difference Test is the confidence level of ESMFold protein structure prediction. This metric is widely used in methods such as AlphaFold2, ESMFold, and OpenFold. pLDDT filters are often used in protein design (such as RFDiffusion), which can significantly improve the success rate of protein design;

\ipara{Natural Sequences Comparisons with Foldseek.} Foldseek takes protein structure as input and searches for proteins with similar structure in the database. We use the experimentally-resolved protein structure as the database (PDB database) to explore how the structure of the generated sequences close to PDB (a higher TM-score indicates higher structural similarity). This method has been used to evaluate other methods for protein sequence generation (ProGen2, ProtGPT2);

\ipara{Diversity Analysis.} We cluster the two sets of sequences (ProGen2-xlarge and CLM) according to sequence similarity. Sequences with a identity higher than 50\% will stay in one cluster. Since the number of input sequences is similar (8,466 vs 8,263), we can measure the diversity of the generated sequences by comparing the number of clusters.

\section{Convergence Analysis of Downstream Fine-tuning Tasks}
\label{app:down} Observing the learning curves in Figure~\ref{fig:scratch_10.7B_vs_esm3b}, we can assess the effectiveness of different fine-tuning scenarios. For the contact prediction task, the convergence speed under the LoRA setting is very similar for both models. Our testing reveals closely matching results for ESM-2 models with capacities of 650M, 3B, 15B, consistent with the findings reported by Ankh et al.~\cite{elnaggar2023ankh}. This similarity suggests possible saturation of the dataset under single-sequence pre-trained models. Additionally, the convergence rates for tasks such as fold classification and fluorescence are generally faster than those for ESM-2, indicating robust generalization following our data augmentation strategies.

Based on the two 470M models defined in our Table~\ref{tab:model_comparison}, despite using the same computational power, we observe distinct outcomes~(Figure~\ref{fig:scratch_vs_transfer_470m}) in contact prediction and fold classification tasks. The MLM model from CLM pre-training converges slightly faster than MLM from scratch. However, the distinction is less pronounced in the two downstream regression tasks. This suggests that perplexity is more sensitive to protein structure related tasks, i.e., contact prediction and fold classification, but shows less sensitivity to regression tasks, particularly when assessed using the Spearman metric, which is prone to variability.

\begin{figure}[!ht]
    \centering
    \begin{subfigure}{0.9\textwidth}
        \centering
        \includegraphics[width=1.0\textwidth]{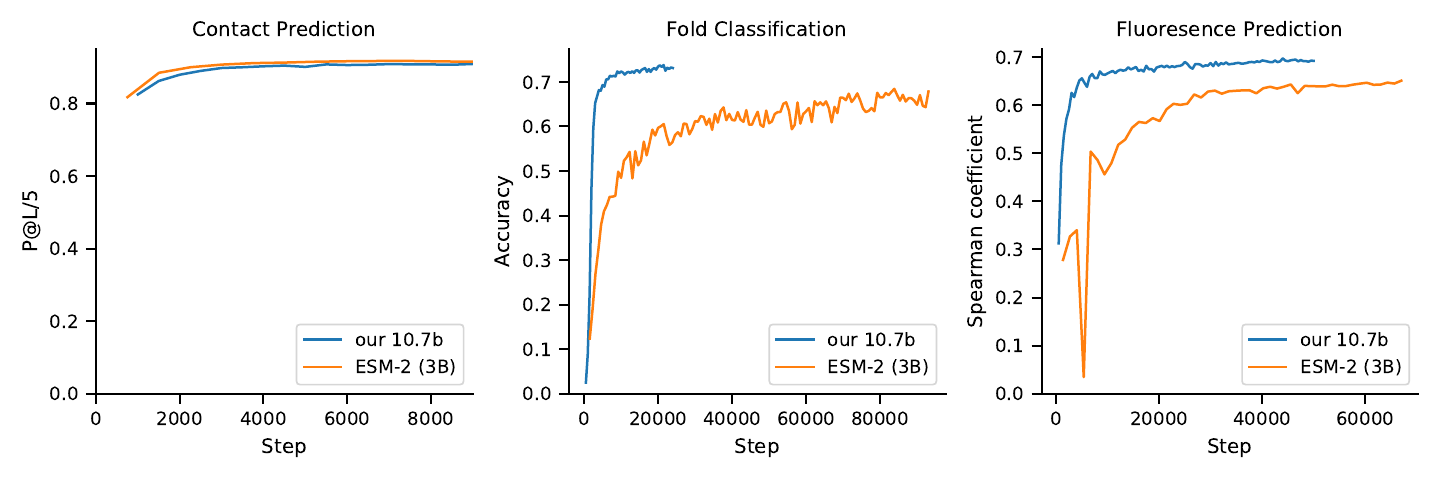}
        \caption{\textbf{Learning Curve Convergence Speed.} LORA fine-tuning our 10.7B model and ESM-2 (3B) model on three downstream tasks.}
        \label{fig:scratch_10.7B_vs_esm3b}
    \end{subfigure}%
    \vspace{1cm} % Adjust the vertical space between the figures as needed
    \begin{subfigure}{0.9\textwidth}
        \centering
        \includegraphics[width=1.0\textwidth]{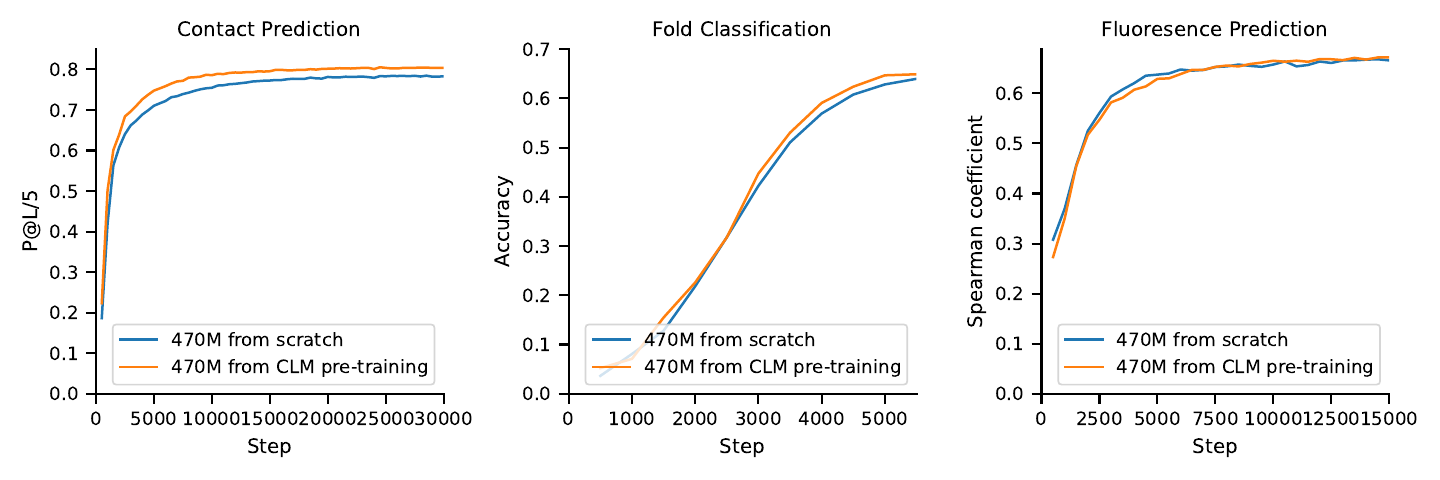}
        \caption{\textbf{Learning Curve Convergence Rate Detection.}. LORA fine-tuning two 470M models on three downstream tasks. \textbf{transfer} means first pre-training 21B tokens on CLM then fine-tuning on MLM with 85B tokens, \textbf{from scratch} means training on 106B tokens from scratch.}
        \label{fig:scratch_vs_transfer_470m}
    \end{subfigure}
\end{figure}

\section{Mixed Objectives Training}
\label{app::mix}

\begin{figure}[ht]
	\centering
	\includegraphics[width=0.99\textwidth]{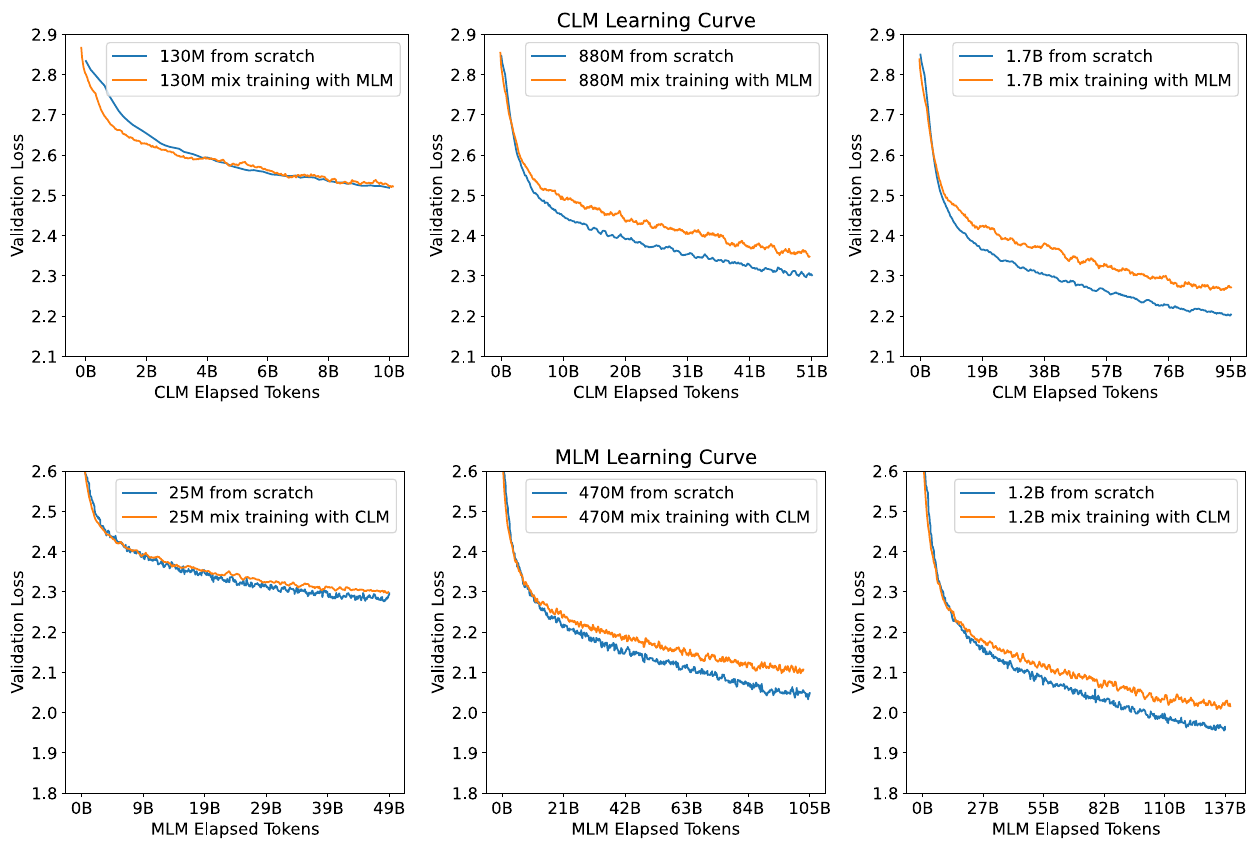}
	\caption{\textbf{Mixed objective validation loss}. Comparative validation loss curves for models trained from scratch versus mixed training approaches. Each panel corresponds to different model sizes, as indicated by the parameters. For each model, two training strategies were compared over an identical number of elapsed tokens: training from scratch (blue) and mixed training with the other objective (orange). Across all model sizes, training from scratch consistently achieves lower validation loss compared to mixed training, suggesting that mixed training may not be as effective as dedicated training for each individual objective.}
  \label{fig::mix_training_curve}

\end{figure}

We also employed an untied model to simultaneously optimize two objectives: 
\[
L_{CLM} = \text{CE}(V\sigma(W_1 ( \text{ encoder}(x))), y_{\text{next}}),
\]
and
\[
L_{MLM} = \text{CE}(V\sigma(W_2(\text{encoder}(x))), y_{\text{mask}}),
\]
where $V$ represents the protein vocabulary embedding, and $W_1$ and $W_2$ are the parameters corresponding to the CLM and MLM tasks, respectively. $\texttt{CE}$ is the cross-entropy operator. The $\sigma$ is the $\texttt{Tanh}$ activation function. 

We compared CLM and MLM under our scaling law of optimal model and data size distributions. One approach involved training from scratch, while the other used mixed training. In the mixed training approach, the actual number of training tokens was higher due to the additional FLOPs consumed by another optimally trained objective, in other words.
In other words, mixed training consumes the FLOPs of two optimal allocations; we only extracted the loss curve of one target for comparison.
We extracted the loss curve of just one target for comparison with the from-scratch training. 
Our findings indicate that mixed training of the two targets can lead to detrimental interference, an effect not observable in smaller models, as depicted in Figure~\ref{fig::mix_training_curve}.
As the model size increases to a hundred million or billion parameters, the differences become more pronounced. 
The possible reason for this situation is that mixed training has reduced the batch size for one of the objectives, making optimization difficult. We did not further investigate the impact of increasing the batch size and only observed based on the training tokens. However, we cannot rule out the possibility that they are mutually detrimental.
Therefore, if both objectives are to be optimized concurrently, a sequential training strategy should be employed: first optimizing CLM, followed by MLM training. We consider that CLM is more challenging to predict than MLM, which may allow the model to capture more complex and implicit sequential features initially, thereby enhancing its ability to understand and predict masked words in subsequent MLM training.

\section{MoE Scaling}
\label{app:moe}
\begin{figure}[!ht]
	\centering
\includegraphics[width=0.99\textwidth]{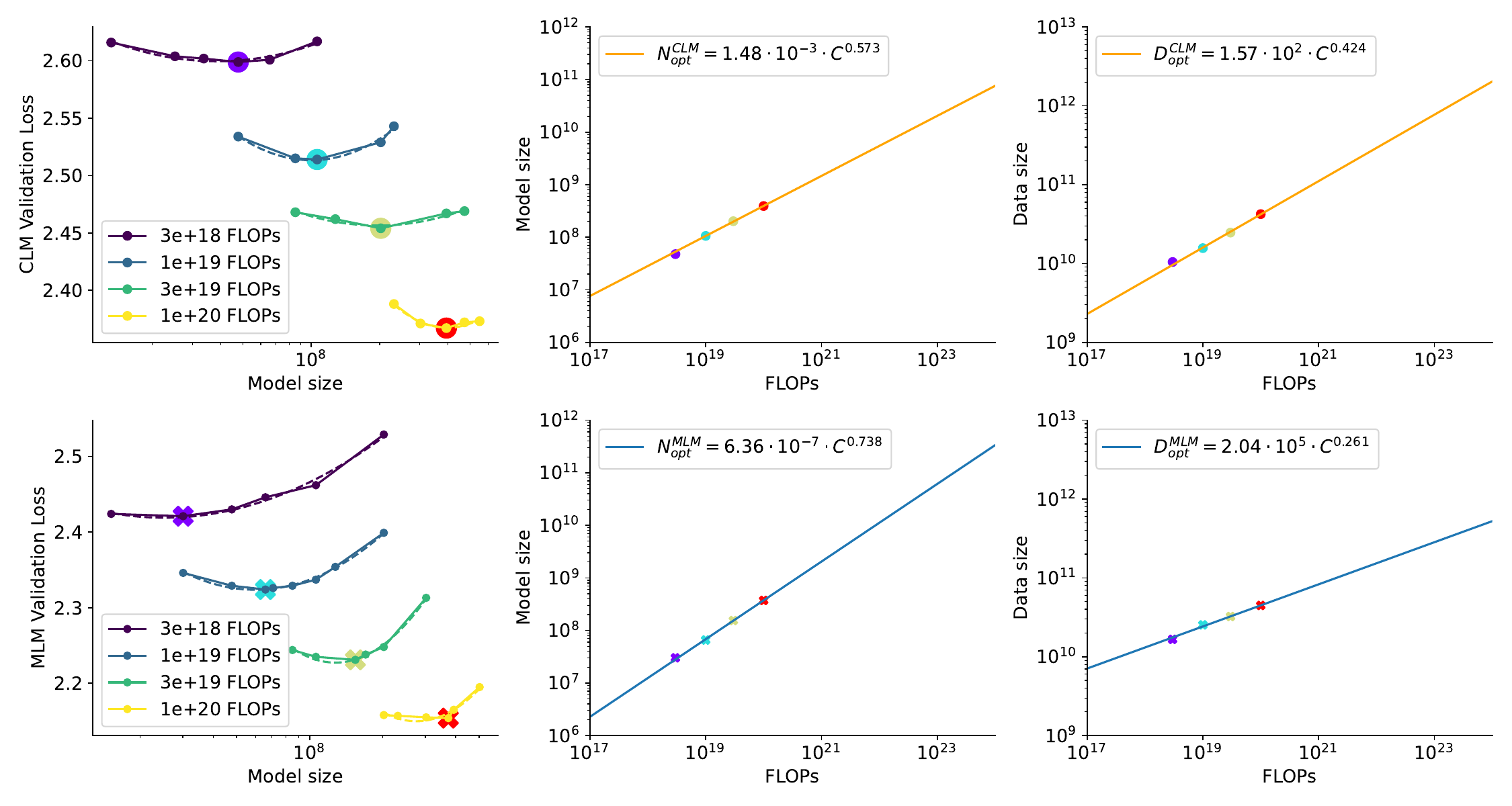}
	\caption{\textbf{Scaling laws of MoE.}The scaling behaviors of sparse parameter counts~(8 experts) in MoE models, highlighting IsoFLOPs curves for different model sizes and FLOPs configurations. Each graph represents the relationship between model size, FLOPs, and validation loss for both CLM and MLM using MoE configurations. The power-law fits indicate optimal model size and data requirements for efficient scaling, showing that MoE models closely align with dense models in terms of scaling efficiency, with power-law coefficients for MoE-CLM and MoE-MLM approximating those of their dense counterparts. This suggests that MoE models can achieve similar scaling behaviors with potentially lower computational costs.} 
%\section{Retrieval-augmented MSA Scaling}
\end{figure}

We find that the scaling behaviors of sparse parameter counts in Mixture of Expert (MoE) models are remarkably similar to those of dense model sizes, potentially allowing for a reduced compute budget for modeling scaling behaviors due to less activated parameters per token. 
%\cxy{The parameters}

In our experiments, we evaluate MoE models ranging from 10M to 500M sparse parameter counts, using a model size of 17 with eight experts, following the settings outlined in Mixtral of experts~\cite{jiang2024mixtral}, including its load-balancing scheme. The figure below shows different IsoFLOPs curves. Notably, the FLOPs here are calculated based on sparse parameters rather than actually activated ones. We use the method described in the main text to select optimal loss points and fit these around the sample points, enabling us to project the optimal model size and number of tokens for larger models (center and right). We observe that the power-law coefficients for CLM and MLM are similar to those of dense models, with MoE CLM vs. Dense CLM at approximately 0.57 vs. 0.58, and MoE MLM vs. Dense MLM at 0.74 vs. 0.77.

Our study strictly focuses on models with eight experts, which may not be entirely rigorous. Clark et al.~\cite{clark2022unified} proposed a unified scaling law defining effective training parameters for MoE, aiming to harmonize the scaling laws for Dense and MoE models. Investigation of biological data will be considered as future work.

%We simply take experts = 8 and run MoE~\cite{jiang2024mixtral} scaling behavior. We find that the sparse parameters have similar trends with our dense Transformer.

%The loss is smaller than in dense models.

%This can be evaluate the dense model with lower compute.

%but it is not strict for different experts number.

%\cxy{unitied}
%\cxy{what is the mix training setting. 
%mix tokens much higher than from scratch.}

%\cxy{how many transferred-clm training.}

\section{Combined Power-law}
\label{app:combine}
We applied the fitting function proposed by Chinchilla~\cite{hoffmann2022training}, detailed in Equation~\ref{eq::chinchilla}, to model the effects of various factors on model performance. 
It can provide a loss prediction where neither the parameters or model size are not optimal allocation.
This loss function simultaneously depends on parameters $N$ and $D$:
\begin{equation}
\label{eq::chinchilla}
L(N, D) = \frac{A}{N^\alpha} + \frac{B}{D^\beta} + E
\end{equation}
where $E$ denotes the irreducible loss. Parameters $A$, $B$, $\alpha$, and $\beta$ are learned through the fitting process. As $N \to \infty$ or $D \to \infty$, the function degenerates to a form similar to Equation~\ref{eq::individual}, which indicates that it models the scenarios under perfect conditions of other variables.

Given that most of our training tokens are used for less than or equal to one epoch, and that the model size is prone to underfitting at fixed FLOPs, the asymptotic behaviors $L(N)$ at $D \to \infty$ and $L(D)$ at $N \to \infty$ are enough for determining the parameters in $L(N, D)$.

To enrich data points, we randomly added several FLOP counts into 25\% of the model size and trained these models for 0.25, 0.5, 0.75, and 1 epoch. And we adopt the Huber loss to fit these coefficients:
\begin{equation}
\min_{a, b, e, \alpha, \beta} \sum_{\text{i}} \text{Huber}_\delta \left( \text{LSE}\left(a - \alpha \log N_i, b - \beta \log D_i, e - \log L_i\right) \right),
\end{equation}
where LSE represents the log-sum-exp operator, and $\delta = 10^{-3}$. The terms $N_i$, $D_i$, and $L_i$ denote the model size, dataset size, and loss of the $i$-th run, respectively.
We fitted the MLM validation loss from 110 samples and the CLM validation loss from 149 samples using grid search with $\alpha \in \{0, 0.5, \ldots, 2\}$, $\beta \in \{0, 0.5, \ldots, 2\}$, $e \in \{-1, -0.5, \ldots, 1\}$, $a \in \{0, 5, \ldots, 25\}$, and $b \in \{0, 5, \ldots, 25\}$. 
The final initialized parameters of CLM and MLM both are $[e,a,b,\alpha,\beta] = [1,5,10,0.5,0.5]$.
We set the maximum number of iterations to 1000, and the two objectives were essentially achieved after 360 iterations.
The exponential powers of learned $a$ and $b$ yielded the coefficients $A$, $B$, which were reported in Table~\ref{tab::chinchilla}. 
\begin{table}[h]
\centering
\caption{Coefficient of Equation~\ref{eq::chinchilla} }
\begin{tabular}{@{}lcccc@{}}
\toprule
Objective &  $A$ & $B$ & $\alpha$  & $\beta$   \\ 
\midrule
%CLM & $2 \times 10^{-3}$     & $3.221$  &  $4.778$   &  $0.041$  &  $0.067$  \\
CLM    & $143.9$  &  $22036.5$   &  $0.367$  &  $0.496$  \\
MLM    & $3.365$ &    $7.569$       &  $0.042$  & $ 0.099 $ \\
%MLM &   $ 1 \times 10^{-4}$       & $3.365$ &    $7.569$       &  $0.042$  & $ 0.099 $ \\
\bottomrule
\end{tabular}
\label{tab::chinchilla}
\end{table}

%2.122563756382614, 143.97531038760354, 22036.528838204948, 0.36658655987525074, 0.4955892608715566

Substituting all learned coefficients into the following Equation from the original Chinchilla paper:
\begin{equation}
\begin{aligned}
N_{\text{opt}}(C) = G \left(\frac{C}{6}\right)^a, \quad D_{\text{opt}}(C) = G^{-1} \left(\frac{C}{6}\right)^b \\
where \quad G = \left(\frac{\alpha A}{\beta B}\right)^{\frac{1}{\alpha+\beta}}, \quad a = \frac{\beta}{\alpha + \beta}, \quad b = \frac{\alpha}{\alpha + \beta}.
\end{aligned}
\label{eq:chinchilla_fit}
\end{equation}
The results closely approximate the trends given in Equations~\ref{eq:c_law} and \ref{eq::individual}, confirming our overall findings. 

\section{IsoLoss}
\label{app::isoloss}

\begin{figure}[ht]
	\centering
	\includegraphics[width=0.99\textwidth]{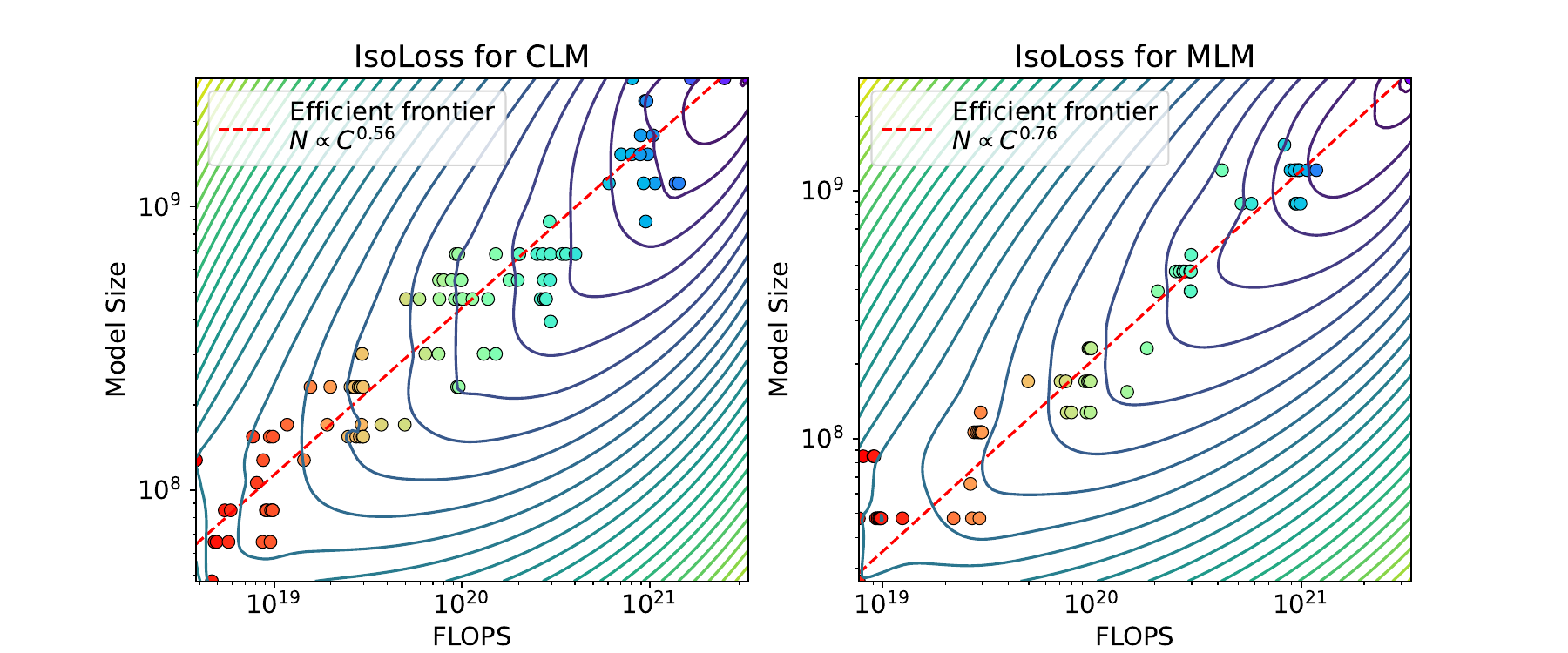}
	\caption{\textbf{Parametric fit for CLM and MLM}. Unlike the IsoFLOPs method used in the main text to select the optimal model size, these plots use all available data points to fit the models. The left panel shows the contour of the function $L$ and the efficient frontier (indicated by the red dashed line) for the CLM, and the right panel for the MLM. The rainbow dots represent identical loss. The results closely align with using the IsoFLOPs profiling method.}
  \label{fig::isoloss}

\end{figure}

In addition to using the seven different FLOPs counts reported in the main text to determine the optimal model sizes and fit our scaling law, we also incorporated additional model points into our analysis. We trained using the final loss points of all the CLM and MLM that are run. Figure~\ref{fig::isoloss} depicts the contour of the fitted function $L$ and the efficient frontier as a red dashed line, presented in log-log space. 
The frontier interval of Figure~\ref{fig::scaling_law} is computed from this observation.
From this approach, it revealed the scaling exponents for model size to be 0.77 in MLM and 0.57 in CLM, very similar to the IsoFLOPs profiling method in Section~\ref{sec::scaling_law}.

\section{Training Procedure}
\label{app:process}
We conducted all experiments using Ampere A100 GPUs (80G) equipped with NVLink, utilizing the GLM framework~\cite{zeng2022glm,du2022glm} developed based on DeepSpeed and Megatron. 
We have used a total of around 1 million GPU hours.
Our approach predominantly utilized data parallelism, avoiding model parallelism and pipeline parallelism to simplify deployment.
Modifications were made to the standard Transformer architecture~\cite{vaswani2017attention}, adopting a DeepNorm~\cite{wang2024deepnet} strategy and layer normalization~\cite{ba2016layer}. The activation function was set to GLU~\cite{shazeer2020glu}, RoPE~\cite{su2024roformer} was used to encode position, similar to the settings found in the Transformer++ architecture~\cite{touvron2023llama1}. 
We further adopt FlashAttention~\cite{dao2022flashattention} to accelerate our training process.
The used max LR empirically found to range between $6 \times 10^{-4}$ and $1.2 \times 10^{-4}$ from small to large model size, was used along with a cosine decay strategy to reduce it to $0.1 \times \text{max LR}$. 
Both CLM and MLM were trained under similar settings for model size, with a consistent LR and a minimum warm-up period of 2.5\% steps, extending to at least 100K training steps.
All sequences were set to a length of 1024, with sequences concatenated using an \texttt{<EOS>} delimiter. Based on findings related to loss magnitude and batch size \cite{mccandlish2018empirical}. 
The AdamW optimizer~\cite{loshchilov2017decoupled} was used with $\beta_1 = 0.9$, $\beta_2 = 0.95, \epsilon=1 \times 10^{-8}$, and a weight decay of $0.01$. 
All experiments omitted the dropout~(it reduced the capacity to hinder model scaling)  and trained with bfloat16.
Most pre-training experiments were confined to the $\leq 1$ epoch, with some models extending up to 30\% beyond one epoch.
For the transfer learning setting,  we load the finished checkpoint of the pre-training model and disregard the pre-trained optimized state, and learn rest tokens with warmup 5\% steps the max LR.

\section{Broader Impact}
\label{app:impact}
If the scaling law of the protein language model improves predictions or understanding of protein structure and function, it could potentially have positive impacts on scientific research in fields such as biology, medicine, and drug development. This may facilitate the development of new drugs, accelerate progress in disease diagnosis, or drive advancements in frontier research in the life sciences.

\section{Model Parameters}
\label{app::modelsize}
Table \ref{table:model_parameters} details the sizes and configurations of all models utilized in this research, training only with data parallel expcept 10B with tensor parallel size 2:
%noting that several models run training repeatedly for varying numbers of steps.

\begin{table}[h]
\centering
\caption{\textbf{All model hyperparameters.} Several of the models presented have been trained using various learning rate schedules and differing amounts of training tokens.}
\begin{tabular}{@{}cccccc@{}}
\toprule
\textbf{params} & \textbf{d\_model} & \textbf{ffw} & \textbf{kv\_size} & \textbf{head\_num} & \textbf{layers} \\
\hline
4M & 192 & 512 & 24 & 8 & 8  \\ 
5M & 256 & 683 & 32 & 8 & 7 \\
6M & 256 & 683 & 32 & 8 & 8 \\
10M & 320 & 853 & 40 & 8 & 8 \\
13M & 320 & 1280 & 40 & 8 & 8 \\
19M & 448 & 1194 & 64 & 7 & 8 \\
25M & 512 & 1365 & 64 & 8 & 8 \\
34M & 512 & 2048 & 64 & 8 & 8 \\
40M & 576 & 1536 & 64 & 8 & 10 \\
47M & 576 & 1536 & 64 & 9 & 12 \\
66M & 640 & 2560 & 64 & 10 & 10 \\
77M & 480 & 1280 & 24 & 20 & 28 \\
85M & 768 & 2048 & 64 & 12 & 12 \\
106M & 768 & 2048 & 48 & 16 & 15 \\
127M & 768 & 2048 & 48 & 16 & 18 \\
154M & 896 & 2389 & 64 & 14 & 16 \\
157M & 640 & 1707 & 32 & 20 & 32 \\
170M & 768 & 2048 & 48 & 16 & 24 \\
200M & 896 & 2389 & 64 & 14 & 21 \\
230M & 896 & 2389 & 64 & 14 & 24 \\
300M & 1024 & 2731 & 64 & 16 & 24 \\
393M & 1280 & 3413 & 80 & 16 & 20 \\
470M & 1280 & 3413 & 80 & 16 & 24 \\
550M & 1280 & 3413 & 80 & 16 & 28 \\
670M & 1536 & 4096 & 96 & 16 & 24 \\
880M & 1792 & 4778 & 64 & 28 & 23 \\
1.2B & 2048 & 5461 & 64 & 32 & 24 \\
1.5B & 2304 & 6144 & 64 & 36 & 24 \\
1.7B & 2304 & 6144 & 64 & 36 & 28 \\
2.0B & 2560 & 6832 & 64 & 40 & 26 \\
2.4B & 2560 & 6832 & 64 & 40 & 30 \\
2.8B & 2560 & 6832 & 64 & 40 & 36 \\
3.1B & 2688 & 7168 & 64 & 42 & 36 \\
3.4B & 2816 & 15040 & 128 & 22 & 22 \\
4.0B & 3072 & 8192 & 128 & 24 & 36 \\
5.7B & 3328 & 8874 & 128 & 26 & 40 \\
6.2B  & 3584 & 9556 & 128 & 28 & 40\\
7.2B & 4096 & 10923 & 128 & 36 & 36 \\
10.7B & 4352 & 11605 & 136 & 32 & 47 \\
\bottomrule
\end{tabular}
\label{table:model_parameters}
\end{table}

\end{document}